\documentclass[pmlr]{jmlr} 

\usepackage{longtable}

\usepackage{booktabs}
\usepackage[]{siunitx}
\usepackage{ulem}

\usepackage[utf8]{inputenc} 
\usepackage[T1]{fontenc}    
\usepackage{hyperref}       
\usepackage{url}            
\usepackage{booktabs}       
\usepackage{amsfonts}       
\usepackage{bbm}
\usepackage{nicefrac}       
\usepackage{microtype}      
\usepackage{xcolor}         
\usepackage{ulem}
\usepackage{rotating} 
\usepackage{multirow}

\usepackage{graphicx}
\usepackage{color}
\usepackage{xcolor}
\usepackage{amssymb}
\PassOptionsToPackage{hyphens}{url}
\usepackage{hyperref}
\usepackage{csquotes}
\usepackage{mdframed}
\usepackage{verbatim}   
\usepackage{ifthen}
\usepackage{lipsum}
\usepackage[symbol*]{footmisc}
\usepackage{microtype}

\theorembodyfont{\upshape}
\theoremheaderfont{\scshape}
\theorempostheader{:}
\theoremsep{\newline}

\jmlrvolume{LEAVE UNSET}
\jmlryear{2023}
\jmlrsubmitted{LEAVE UNSET}
\jmlrpublished{LEAVE UNSET}
\jmlrworkshop{NeurIPS 2022 -- Traffic4cast} 

\def\eg{\textit{e.\,g.}}
\def\ie{\textit{i.\,e.}}

\def\t4c{\textit{Traffic4cast}\relax}

\let\cite\citep

\newcommand{\showtodos}{show them}  
        {
            \ifthenelse{\isundefined{\showtodos}}%
                    {\expandafter\comment}%
                    {\begin{mdframed} \textcolor{red}{{\bf TODO}}}%
                    }%
         {%
            \ifthenelse{\isundefined{\showtodos}}%
                    {\expandafter\endcomment}%
                    {\end{mdframed}}%
          }

\def\eg{\textit{e.\,g. }}
\def\ie{\textit{i.\,e. }}

\def\t4c{\textit{Traffic4cast}\relax}

\def\m9#1{\textcolor{red}{\textsf{\quad[[m9: #1]]}}}

\title[\t4c at NeurIPS 2022]{\t4c at NeurIPS 2022 --
Predict Dynamics along Graph 
Edges from Sparse Node Data:
Whole City Traffic and ETA from Stationary Vehicle Detectors
\\ {\small \url{http://traffic4cast.ai} -- \url{https://github.com/iarai/NeurIPS2022-traffic4cast}}
}

\newcommand\blfootnote[1]{%
  \begingroup
  \renewcommand\thefootnote{}\footnote{#1}%
  \addtocounter{footnote}{-1}%
  \endgroup
}

\author{
\Name{Moritz Neun}\thanks{Institute of Advanced Research in Artificial Intelligence (IARAI), Vienna, Austria}$^{,a}$ \Email{moritz.neun@iarai.ac.at}
\AND
\Name{Christian Eichenberger}\footnotemark[1]$^{,a}$ \Email{christian.eichenberger@iarai.ac.at}
\AND
\Name{Henry Martin}\footnotemark[1]$^,$\thanks{Institute of Cartography and Geoinformation, ETH Zurich, Switzerland} \Email{henry.martin@iarai.ac.at}
\AND
\Name{Markus Spanring}\footnotemark[1] \Email{markus.spanring@iarai.ac.at}
\AND
\Name{Rahul Siripurapu}\footnotemark[1] \Email{rahul.siripurapu@iarai.ac.at}
\AND
\Name{Daniel Springer}\footnotemark[1] \Email{daniel.springer@iarai.ac.at}
\AND
\Name{Leyan Deng}\thanks{School of Data Science, University of Science and Technology of China}$^,$\thanks{Huawei Noah’s Ark Lab},
\Name{Chenwang Wu}\footnotemark[3],
\Name{Defu Lian}\footnotemark[3],
\Name{Min Zhou}\footnotemark[4]
\Email{dleyan@mail.ustc.edu.cn}
\AND
\Name{Martin Lumiste}\thanks{Bolt Technology, Tallinn, Estonia},
\Name{Andrei Ilie}\footnotemark[5]$^,$\thanks{University of Bucharest, Bucharest, Romania}
\Email{martin.lumiste@bolt.eu}
\AND
\Name{Xinhua Wu}\thanks{Department of Civil and Environmental Engineering Northeastern University Boston, MA, USA},
\Name{Cheng Lyu}\thanks{Chair of Transportation Systems Engineering, Technical University of Munich, Germany},
\Name{Qing-Long Lu}\footnotemark[8],\\\mbox{      }\Name{Vishal Mahajan}\footnotemark[8]
\Email{wu.xinh@northeastern.edu}
\AND
\Name{Yichao Lu}\thanks{Layer 6 AI, Toronto, Canada} \Email{yichao@layer6.ai}
\AND
\Name{Jiezhang Li}\thanks{School of Coumpute Science and Engineering, South China University of Technology, Guangzhou, China},
\Name{Junjun Li}\footnotemark[10],
\Name{Yue-Jiao Gong}\footnotemark[10]
\Email{gongyuejiao@gmail.com}
\AND
\Name{Florian Grötschla}\thanks{ETH Zurich, Switzerland},
\Name{Joël Mathys}\footnotemark[11]
\Email{fgroetschla@ethz.ch}
\AND
\Name{Ye Wei}\thanks{Department of Computer Science Loughborough University Loughborough, UK},
\Name{He Haitao}\thanks{School of Architecture, Building and Civil Engineering Loughborough University Loughborough, UK},
\Name{Hui Fang}\footnotemark[12]
\Email{y.wei@lboro.ac.uk}
\AND
\Name{Kevin Malm}\thanks{HERE Technologies, Chicago, IL, USA}\Email{kevin.malm@here.com}
\AND
\Name{Fei Tang}\thanks{Kaiko, Zurich, Switzerland}  \Email{fei.tang3@gmail.com}
\AND
\Name{Michael Kopp}\footnotemark[1] \Email{michael.kopp@iarai.ac.at}
\AND
\Name{David Kreil}\footnotemark[1] \Email{david.kreil@iarai.ac.at}
\AND
\Name{Sepp Hochreiter}\thanks{Machine Learning Institute, Johannes Kepler University Linz, Austria}$^,$\footnotemark[1] \Email{sepp.hochreiter@iarai.ac.at}
\AND
}

\begin{document}
\maketitle
\addtolength{\footnotesep}{3mm}
\blfootnote{\hspace{-0.25cm}$^a$ Equal Contribution}

\clearpage

\begin{abstract}

The global trends of urbanization and increased personal mobility force us to rethink the way we live and use urban space.
The \t4c competition series tackles this problem in a data-driven way, advancing the latest methods in machine learning for modeling complex spatial systems over time.
In this edition, our dynamic road graph data combine information from road maps, $10^{12}$ probe data points, and stationary vehicle detectors in three cities over the span of two years.
While stationary vehicle detectors are the most accurate way to capture traffic volume, they are only available in few locations.

\t4c 2022 explores models that have the ability to generalize loosely related temporal vertex data on just a few nodes to predict dynamic future traffic states on the edges of the entire road graph.

In the core challenge, participants are invited to predict the likelihoods of three congestion classes derived from the speed levels in the GPS data for the entire road graph in  three cities \SI{15}{\minute} into the future.
We only provide vehicle count data from spatially sparse stationary vehicle detectors in these three cities as model input for this task.
The data are aggregated in \SI{15}{\minute} time bins for one hour prior to the prediction time.
For the extended challenge, participants are tasked to predict the  average travel times on super-segments \SI{15}{\minute} into the future -- super-segments are longer sequences of road segments in the graph.

The competition results provide an important advance in the prediction of complex city-wide traffic states just from publicly available sparse vehicle data and without
the need for large amounts of real-time floating vehicle data.

\end{abstract}

\section{Introduction}

\begin{figure}[htb]
  \centering
  \includegraphics[width=0.85\linewidth]{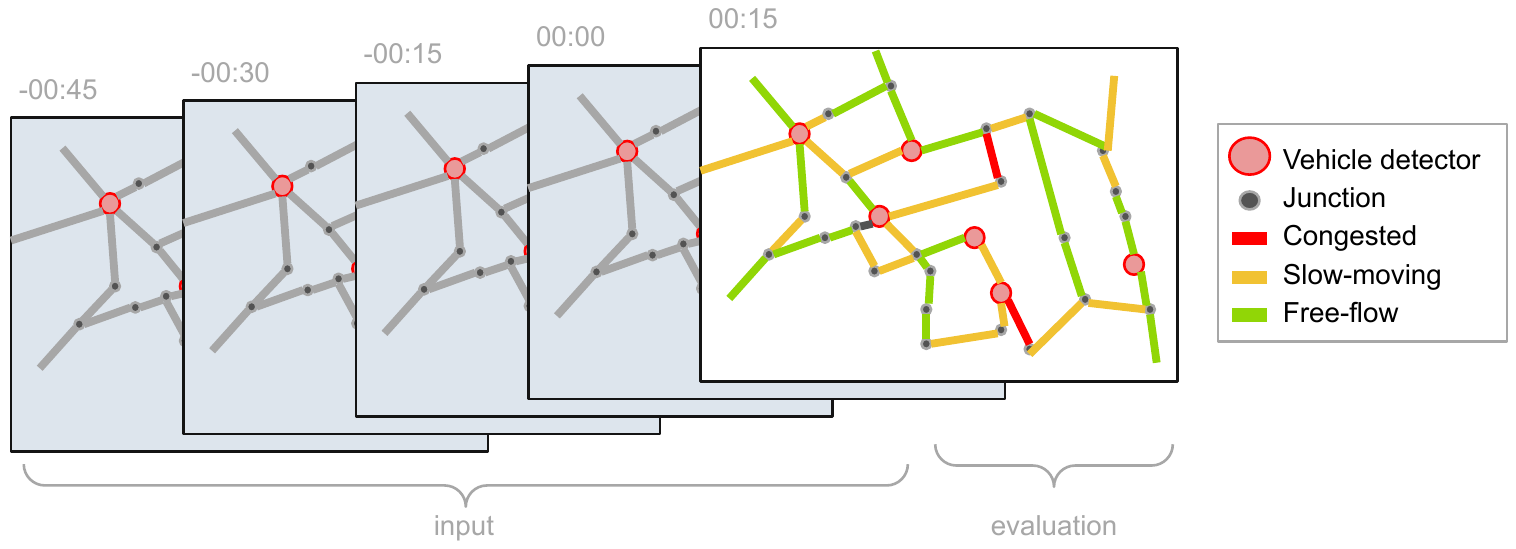}
  \vspace*{-5mm}
  \caption{\t4c 2022: Traffic Dynamics Prediction on
Graphs for the Congestion Classification Task. Given one hour of sparse vehicle count data only (red disks) the task is to predict the congestion classification for all road segments \SI{15}{\minute} into the future (green, yellow, and red lines). 
For the extended challenge, the prediction targets are average travel times along super-segments.} 
  \label{fig:system_evolution_new}
  \vspace*{-5mm}
\end{figure}

Going beyond the \t4c challenges at NeurIPS 2019, 2020, and 2021 \citep{Kreil_Traffic4cast_2019,Kopp_Traffic4cast_2020,eichenberger_traffic4cast_2021}, \t4c 2022 explores models that have the ability to generalize loosely related temporal vertex data on just a few nodes to predict dynamic future traffic states on the edges of the entire road graph.
In our {\bf core challenge}, participants are asked to predict the likelihood of three congestion classes derived from for three cities for the entire road graph \SI{15}{\minute} into the future. A schematic overview is shown in Figure~\ref{fig:system_evolution_new}, illustrating the common red, yellow, or green coloring of roads on a traffic map.
We provide vehicle count data from spatially sparse stationary vehicle detectors in these three cities in \SI{15}{\minute} aggregated time bins for one hour prior to the prediction time slot.
Stationary vehicle detectors are often inductive loops (electrically conducting loops) installed in the road pavement \cite{wikipedia_induction_loop} or cameras at traffic lights, and providing vehicle counts and sometimes also speed and occupancy data -- \t4c 2022 only uses the vehicle count data.
See Figure~\ref{fig:three_cities_with_graph_new} for the location of these stationary vehicle detectors in relation to the rest of the road graph. For our {\bf extended challenge}, participants are tasked to predict the average travel times on super-segments \SI{15}{\minute} into the future. Average travel times correspond to ETAs (Expected Time of Arrival), and we loosely use both terms synonymously. Super-segments are longer sequences of road segments in the graph.

Solving these challenges has direct substantial implications for our ability to forecast, plan, and analyze urban traffic leading to a considerable impact on society, environment, and health.
Our competition addresses the two main challenges identified in \citep{manibardo_deep_2020} for the application of deep
learning methods for road traffic forecasting by formulating the task in an actionable (\ie application-driven) setting and by providing the  dataset and leaderboard.
Its challenges directly allow for ETA (estimated time of arrival) prediction (see \citep{hu_deepeta_2022,schleibaum_explainable_2022}) from publicly available sparse data.
Therefore, without the need for large amounts of real-time floating vehicle data, it dramatically lowers the technical and financial entry barriers for performing such predictions. 
This is an active area of research \citep{derrow-pinion_eta_2021, hu_deepeta_2022,elmasri_beyond_2019,schleibaum_explainable_2022}, which our competition may directly impact.
The vast city-wide scope of our competition's underlying data directly boosts the active field of research on full-field traffic state identification. Currently, the traffic state identification is often limited to particular road sub-systems for which data is available, \eg \citep{zhang_hybrid_2022}.

Moreover, the ability to generally predict temporal edge features in graphs from sparse vertex data is crucial for many real-world applications, many of which are critical to the functioning of our society.
In \cite{https://doi.org/10.48550/arxiv.2007.03113} the spread of COVID-19 can be modelled as predicting temporal edge features in a large spatio-temporal graph.
Malicious software can be identified as anomalous patterns in temporal sequences of evolving multiscaled API graph \citep{https://doi.org/10.1002/int.22880}.
Solutions to traffic prediction on graphs are also relevant to Wide Area Network traffic prediction \citep{WAN} and to learning the temporal dynamics of crypto-currency networks \citep{https://doi.org/10.48550/arxiv.1902.10191}. 

\subsection{Data}
We provide a unique data set derived from massive industry-scale GPS data, large openly available stationary vehicle detector data, and road graph information from three diverse metropolitan areas.

\begin{figure}[htb]
  \centering
  \includegraphics[width=0.7\linewidth]{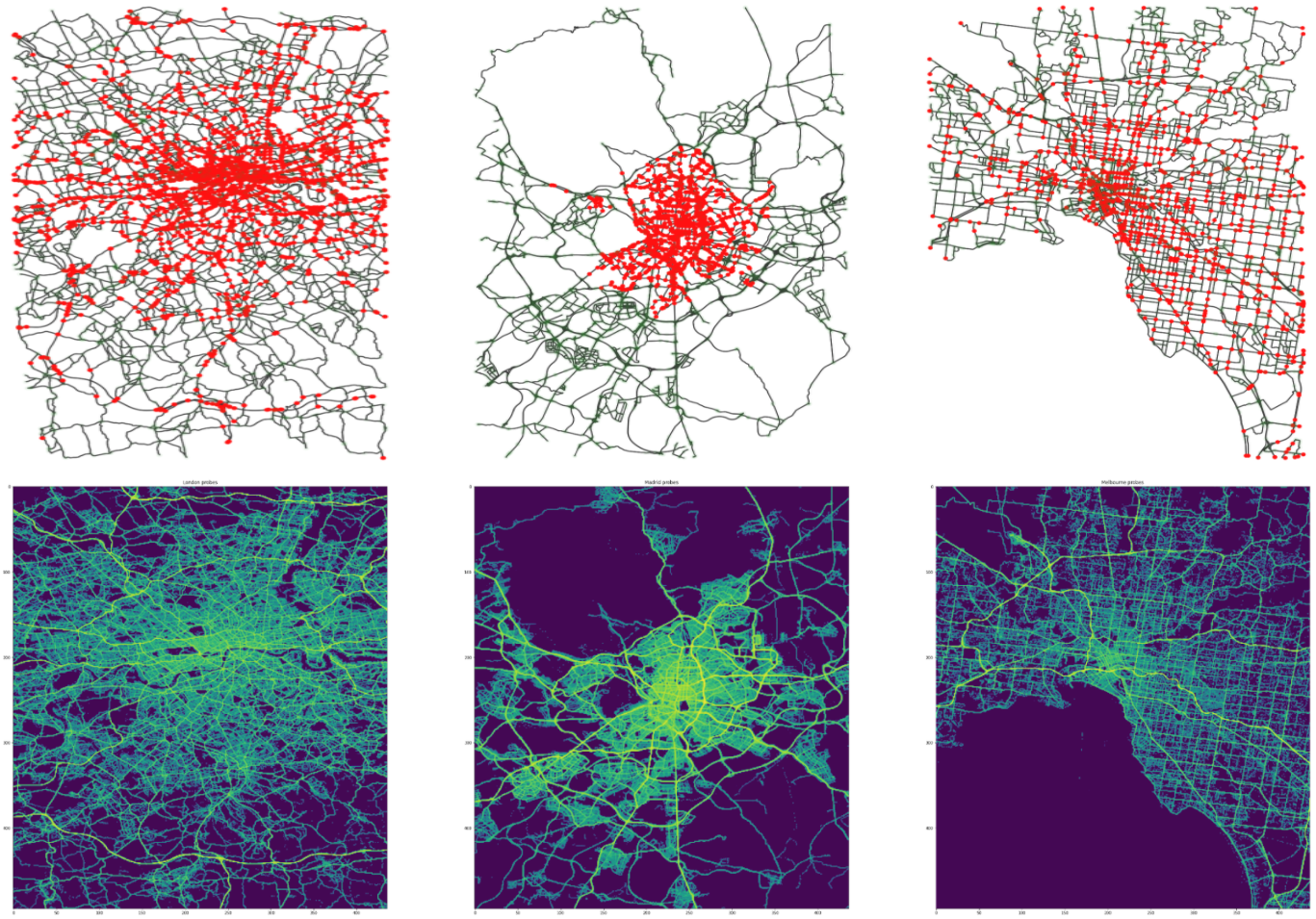}
  \vspace*{-4mm}
  \caption{From left to right we show London, Madrid, and Melbourne.
  Top panels: distribution of stationary vehicle detector locations (red dots) in comparison to the sampled road networks.
  Bottom panels: GPS probe data covering a much finer and larger road network. The road graph sampling removed road segments with very few data points.
  Sources \copyright HERE Technologies, \copyright OpenStreetMap Contributors, \copyright TfL and Highways England, \copyright Madrid City Council, \copyright Victoria State Government
  } 
  \label{fig:three_cities_with_graph_new}
  \vspace*{-4mm}
\end{figure}

The GPS data has been aggregated and made available 
by HERE Technologies \cite{here:sample-data}
as in the previous instances of \t4c \citep{Kreil_Traffic4cast_2019,Kopp_Traffic4cast_2020,eichenberger_traffic4cast_2021}. The data originates from a large fleet of probe vehicles which recorded their movements in the years 2019 to 2021.
For this year's competition we focus on three cities: London, Madrid, and Melbourne (see Figure~\ref{fig:three_cities_with_graph_new}).
The underlying raw data is similar to the input for commercial traffic maps and routing products.
Unlike previous years, this data is not available as test input, but serves to derive the ground truth labels for our core and extended challenge, see \cite{mets-10} and Appendix~\ref{app:output_labels}.
In addition, participants can use the speed data on the graph edges for jointly modeling traffic or as labels for intermediate tasks. 
A selected set of time intervals for all three cities are held back for the test set. Specifically, we split on alternate weeks into training data and data held back for sampling 100 test slots.

Stationary vehicle detector data is the new additional data source for \t4c 2022.
Stationary vehicle detectors are often made publicly available by city or motorway authorities. Furthermore, an increasing collection of crowd-sourced data through alternative sensor networks \citep{telraam_2021, wecount_2022} is becoming widely available.
Stationary vehicle detectors are spatially sparse (up to $\sim4000$ per city, see Table~\ref{tab:loop_counters_new} and Figure~\ref{fig:three_cities_with_graph_new}) but capture all traffic at their locations.
%
We focus on the large, publicly available, stationary vehicle detector data for London, Madrid, and Melbourne.
The stationary vehicle detector measurements are represented as nodes with measured volumes per \SI{15}{\minute} time bin.

We are providing a pre-compiled version of the derived data sets as well as tools to convert or easily load the data in our GitHub repository.

\subsubsection{Input and Static Data}

The input data for our core and the extended challenge is derived from publicly available stationary vehicle detector data sets from Madrid, Melbourne, and London (see Table~\ref{tab:loop_counters_new}).
See Appendix~\ref{app:loop_counters} for more details.
The stationary vehicle detector measurements can be attached to the nodes in the provided road graph (see also Figure~\ref{fig:system_evolution_new}).

\begin{table}[hbt!]
    \centering
    \begin{tabular}{|l|p{2.3cm}|p{2.cm}|p{1.6cm}|p{1.6cm}|p{1.5cm}|}
    \hline
          city & \# vehicle detectors  / \# nodes with detector data  
          & \# nodes  (incl. detectors) & \# edges & \# super-segments  \\ \hline\hline
         London \citep{london_tims_tfl,london_highways_england} & 3751 / 3751
                & \hfill 59110& \hfill  132414 & \hfill 4012\\ \hline
         Madrid \citep{madrid_historico_de_datos_de_trafico_desde_2013} &  3840  / 3875
            & \hfill  63397 & \hfill  121902 & \hfill 3969\\ \hline
         Melbourne \citep{melbourne_traffic_signal_volume_data} & 2589 / 3982
            & \hfill  49510 & \hfill  94871 & \hfill 3246 \\ \hline
    \end{tabular}
    \caption{Number of stationary vehicle detectors and road graph nodes, edges, and super-segments  for the three cities London, Madrid, and Melbourne. Vehicle detector data may be split across several nearby nodes.}
    \label{tab:loop_counters_new}
    \vspace*{-2mm}
\end{table}

We are providing a road graph, represented by a directed graph, for each city with stationary vehicle detector data.
The road graph is derived from OpenStreetMap \citep{OpenStreetMap2022}, simplified by filtering out local roads with no or almost no GPS ground truth and collapsing nearby nodes that are not relevant for the traffic flow and connectivity. See Table~\ref{tab:loop_counters_new} for an overview of the graph vertex and edge sizes and Appendix~\ref{app:road_graph} for more details on the derivation.

In addition, we are providing a choice of super-segments \citep{derrow-pinion_eta_2021} represented as lists of segment IDs.
The use of super-segments is motivated by contraction hierarchies \citep{geisberger_2008}, which are a common strategy in most commercial routing engines.
Super-segments are sampled from the road graph and validated using a commercial routing engine. See Appendix~\ref{app:super_segments} for details. 

\subsubsection{Output Labels}
The aggregated GPS probe data are used to generate our ground truth labels.
As the stationary vehicle detector data comes at \SI{15}{\minute} intervals, the \SI{5}{\minute} time bins in the spatio-temporal data format that has been used in previous editions of the competition, see \citep{eichenberger_traffic4cast_2021}, are aggregated to \SI{15}{\minute} time bins \cite{mets-10}.
We then derive the following ground truth labels:
\begin{itemize}
    \item Congestion Class (\textbf{CC}; red/congested, yellow/warning, green/uncongested) for each segment in the road graph.
    The class is derived from the aggregated GPS probe data; if not enough data is available to derive the congestion class, it will output missing value.
    The code to derive the labels from the speed data is available to participants.
    \item Travel Time (\textbf{ETA}) for each super-segment.
    This is calculated using the the speeds in the segments in the super-segment weighted by their length.
    The code to derive the labels from the speed data is available to participants in our GitHub repository.
\end{itemize}
More details can be found in \cite{mets-10} and Appendix~\ref{app:output_labels}.

\subsection{Metrics}\label{sec:metric}


In order to give more weight to the less frequent but nonetheless important red class, we use weighted masked cross-entropy loss \cite{torch.nn.CrossEntropyLoss} on congestion classes. The class weights are derived through macro-averaging \cite{scikit-learn_model_evaluation} the labels in the training set for each city separately and published in our GitHub repository. Missing congestion class values are masked out.
The overall score is computed as the average of the 3 city scores. 
%
%
%
The metric is formally defined in Appendix~\ref{app:metrics}.

We use L1 Loss for the extended competition as a simple domain-specific evaluation score known in the ETA literature \cite{elmasri_beyond_2019, hu_deepeta_2022, derrow-pinion_eta_2021, schleibaum_explainable_2022}.
Notice that in contrast to the core competitions, there is no ``evaluation mask'', \ie{} all super-segments have an ETA label. As in the core competition, we take the average of the 3 city L1 losses.
In addition, sampling super-segments from important road junctions results in oversampling areas with more data, potentially allowing for more accurate predictions than learning historic defaults in situations with temporally and spatially sparse data.

\section{Standout Solutions}

\t4c 2022 saw over 80 submissions. Here, we briefly describe the top contributions.

\subsection{ustc-gobbler: Transposed Variational Auto-Encoder and Graph-Attention Networks}
\citet{ustc-gobbler} use a variational auto-encoder to reconstruct missing vehicle count data for nodes without a stationary vehicle detector on the transposed matrix to get different values at different nodes.
They engineered a lookup of exact time information from the stationary vehicle detector data as stationary vehicle detector data is publicly available.
They embed static edge features to get weights of a GATv2  layer \cite{gatv2}, which is applied on node embeddings and the reconstructed stationary vehicle detector data. 
In addition to the pairs of node features from node embedding and stationary vehicle detector data, embeddings of a k-means volume cluster, time information, and edge index are fed into final dense layers to produce the final congestion classes.
In the extended competition, additionally, the super-segment-edge adjacency matrix is applied on the intermediate edge features before feeding into the final MLPs. 
They use the sum of reconstruction and cross-entropy (resp., L1 loss in the extended competition) as loss function.

\subsection{Bolt: LightGBM leveraging PCA-based feature extraction}
\cite{Bolt} use PCA-based feature extraction to encode the global traffic state and as proxy for time information.
In addition to PCA-based features, they use further city global features, feed target-encoded features, static edge/super-segment features and positional features into a GBM model both for the core and extended competition.
They also use target encodings as score initialization.
Global features dominate the predictions, closely followed by positional and target encodings.

\subsection{TSE: LightGBM with Similarity-Based Feature Extraction }
\citet{tse} use a k-NN approach to derive super-segment features from the input vehicle count data in the extended competition.
The similarity-based feature extraction is inspired by the assumption that vehicle count data are good encoders of traffic state.  
Multiple statistics and multiple neighbor sets are used. Manhattan distance is found to perform best.
A Gaussian process-based approach is used to impute temporally missing data at stationary vehicle detector locations at two periods, both going into the similarity-based features.
In addition to similarity-based features, static super-segment features (incl. historical features) and volume-based features from the underlying stationary vehicle detectors, as well as additional combined features are fed into a GBM.

\subsection{oahciy: Two-Stage GBM (XGBoost and LightGBM)}
\citet{oahciy} passes all 4 dynamic vehicle count data of all nodes to a first-stage XGBoost and LightGBM ensemble to extract month, day of the week, and time of day.
Target encodings are computed at three levels: for time of day, weekday vs. weekend, and day of the week.
The time information is passed to the second-stage XGBoost and LightGBM ensemble along with static features (incl. edge and node indices), dynamic features, and smoothed target encodings at all three levels. 
In the extended competition, the extracted time information is passed along with smoothed target encodings (the dynamic vehicle count information is not used at all).
The target encoding features achieve the highest feature importance scores in both competitions; in the extended competition, the super-segment ID is particularly important as well.

\subsection{GongLab: Multi-Task Learning GNN}
\citet{Gonglab} use embeddings of the dynamic vehicle count data as node features and embedding of static edge properties and a global volume-based target encoding as edge features to go into a GNN.
After a final embedding of node features, edge labels are derived from the pair of neighboring node features.
They not only derive the logits for the congestion classes but also use the speed labels provided in the competition and the volume class from the underlying GPS data in a multi-task setting.
The loss function is a weighted sum of the three corresponding losses.
The labels in the extended competition are derived by summing up the quotients of edge length and edge speed in the super-segments without additional training.

\subsection{discovery: Hierarchical GNN}
\citet{discovery} introduce additional nodes to reflect hierarchical graph structures: 
in the core competition, they add edges between supernodes (important road junctions that were used in the extended competition to derive super-segments), to support the information exchange between ``important'' intersections in dedicated interleaved layers; in the extended competition, additional super-segment nodes are added and connected to all the nodes of the corresponding super-segment, allowing for information pooling from all nodes lying on the same super-segment and then propagating it back to them in dedicated layers.
In the core competition, congestion classes on edges are derived from the two neigboring node features.
In the extended competition, they use the signal both from the congestion classes as well as the ETAs on super-segment nodes for training.
\citet{discovery} use static edge features; in addition to the dynamic vehicle count data, node features contain positional embedding global mean and standard deviation of the vehicle counts.
This combination allows to learn the global traffic state as well as to cover up for spatial and temporal sparsity.

\subsection{ywei: Two-Stream GNN}

\citet{ywei} base their solution on \textit{LinkX} \citep{LINKX}, first extracting node features from city topology (adjacency matrix) and from stationary vehicle detector data separately, and then combining the two to get node features from both sources.
The node features of neighboring nodes are then concatenated and fed into an MLP to get the final congestion prediction on edges.
Surprisingly, this approach does not use any feature engineering nor even static road information at all, only leveraging the road graph information and the dynamic stationary vehicle detector data as input. 
Using MLP-based embedding of node features and adjacency matrix, this approach is in principle able to learn global traffic state information as well as to cover up for spatial and temporal sparsity.

\section{Synopsis and Discussion}

Looking at the different solutions above, we see both Gradient Boosting Methods and GNN approaches.
All participants used models trained on each city separately.
We also see participants experiment with multi-task settings, trying to exploit feedback from both tasks, approaches avoiding feature engineering, and multi-level graphs, to name just a few.
Table~\ref{tab:synopsis} highlights the key aspects and differences of the chosen architectures and informs the discussion below.

\begin{sidewaystable}
    \scriptsize\centering
\begin{tabular}{|p{2.0cm}|p{0.7cm}|p{1cm}|p{1cm}|p{1cm}|p{1.0cm}|p{0.7cm}|p{0.7cm}|p{0.7cm}|p{0.7cm}|p{1cm}|p{1cm}|p{1cm}|p{1cm}|}
\toprule
 Team, rank (c./e.), approach   & city spec.$^a$ & time information/ global state$^b$               & ensemble (p.city) (c./e.)$^c$ & avg. model size (c./e.)$^d$     & total model size (all cities) (c./e.)$^e$& input dim edges (c./e.)$^f$ & input dim super-segments (c./e.) $^g$& input dim nodes (c./e.)$^h$ & static road in\-for\-ma\-tion$^i$ & target encoding (historic dis\-tri\-bu\-tion)$^j$ & node index or position $^k$ & index or position (edges / super\-seg\-ments) $^l$ & edge / super-segment label de\-ri \-va\-tion$^m$
 \\
 \midrule
\textbf{ustc-gobbler (1/1)}\newline  GNN (GATv2\-Conv)  \cite{ustc-gobbler}      & yes        & lookup  & 18.3 / 10.0         & 37.1M / 41.2M  p. & 2.0G / 1.2 G p.           & 10 / 8           & --/ 1 + |E| + |V| & 4/4               & yes                     & no                                      & yes                    & yes                                 & MLP  
\\ \hline
\textbf{Bolt (2/4)} \newline GBM \cite{Bolt}                  & yes        & PCA                                          & 1/1               & 23.5M / 3.41M n.   & 70.6M / 10.2M n.           & 38.0 / --       & --/32.7& --              & yes                     & yes                                     & no                     & yes                                 & --                                               
\\ \hline
\textbf{TSE (6/2)}\newline GBM \cite{tse}                 & yes        & kNN similarity                               & --/2.3            & -- / 249k  n.      & -- / 1.74M n.              & -- / --        & -- /196.4 & --              & yes                     & yes                                     & --                     & yes                                 & --                                               
\\\hline
\textbf{oahciy (3/3)}\newline GBM \cite{oahciy}                 & yes        & two-step GBM                                 & 1/1.7 (*)         & 18.6M / 2.78 M n.  & 55.7M / 13.9M n.           & 59.0 / --       & --/11.7  & --              & yes                     & yes                                     & yes                    & yes                                 & --                                               
\\\hline
\textbf{GongLab (4/6)}\newline GNN \cite{Gonglab}                   & yes        & global volume cluster                        & 9=9               & 479M = 479M p.    & 12.9G = 12.9G p.          & 5=5             & --  & 4=4             & yes                     & yes (lookup based on volume cluster)& no                     & no                                  & Link\-Predictor / sum from predicted speeds        
\\\hline
\textbf{discovery (7/5)}\newline GNN (hierarchical)  \cite{discovery}  & yes        & global volume mean/std (4 x \SI{15}{\minute})           & 1/1               & 6.8M / 6.8M p.    & 20.3M / 20.9M p.          & 14/14         & --/--    & 4/4             & yes                     & no                                      & yes                     & no                                 & Link\-Predictor / pooling edges                    
\\\hline
\textbf{ywei (7/--)}\newline GNN (LinkX) \cite{ywei}         & yes        & MLP                                          & 1/--              & 15.4M / -- p.     & 46.2M / -- p.             & |E|/--              & --/--   & 4/--            & no                      & no                                      & no                     & no                                  & Link\-Predictor / --                               
 \\ \bottomrule
\end{tabular}
    \vspace*{-4mm}
    \caption{Synopsis.
    $^a$ Was the model trained per city?
    $^b$ How does global state information enter the model?
    $^c$ How many trained models per city?
    $^d$ Average number of parameters (non-GBM) resp. number of nodes (GBM) per trained model.
    $^e$ Sum of number of parameters (non-GBM) resp. number of nodes (GBM) of all trained models of all cities.
    $^f$/$^g$/$^h$ number of features per edge/super-segment/node.
    $^i$ Were static road attributes?
    $^j$ Was target encoding used?
    $^k$/$^l$/$^m$ was the node/edge/super-segment index or position used as feature?    
      }
    \label{tab:synopsis}
    \vspace*{-2mm}
\end{sidewaystable}

\subsection{How Did Approaches Deal with Global State/Time Information?}

As we were interested in extracting traffic state from the sparse vehicle detector data alone,  we did not explicitly provide time information in the test set, whereas this kind of information would be available in a production setting.
Many of the participants report the high relative importance of these features and found many different ways  of retrieving such global state such as the time information, often combined with target encoding (see Table~\ref{tab:synopsis}).
\cite{ustc-gobbler} engineered a lookup of time information as stationary vehicle detector data is publicly available.
\cite{Bolt} used PCA to derive global features which identify traffic state well.
\cite{tse} used kNN filters on stationary vehicle detector data to extract similarity-based edge features from the neighbor set.
\cite{oahciy} used a first GBM to extract time information from the dynamic stationary vehicle detector data.
\cite{Gonglab} used global volume clusters to look up target encodings.
\cite{discovery} feed global volume means and standard deviations from the past hour into their model.
\cite{ywei} use an MLP to embed stationary vehicle detector data.

\subsection{Dealing with Temporal and Spatial Sparsity in the Input Data}

The input data was sparse, both temporally and spatially. 
Spatial sparsity means that only \SIrange{6}{8}{\%} of nodes (see Table~\ref{tab:loop_counters_new}) in the road graph have stationary vehicle detector installed, and  temporal sparsity means missing vehicle count values at stationary vehicle detector locations.
The output labels still need to be predicted on the edges of the full graph.
Temporal coverage varies significantly between cities and road types, see \cite{mets-10}.

Tabular GBM approaches are a way to deal with spatial sparsity with appropriate features~\cite{Bolt,tse,oahciy}. All three approaches use target encodings.
On the other hand, many different techniques were used in GNN approaches:
\cite{ustc-gobbler} uses TVAE and node embeddings, which can cover up for temporal and spatial sparsity; \cite{Gonglab} used global volume clusters to lookup target encodings; \cite{discovery} use node positional encoding and global features, which can cover up for temporal and spatial sparsity; \cite{ywei} use an MLP to embed stationary vehicle detector data.

\subsection{Where Was it Hard to Predict?}

\begin{figure}[htbp]
\floatconts
{fig:leaderboard}
{
\vspace*{-8mm}
\caption{Losses in the \t4c 2022 in the leaderboard  for the first three teams for core and extended competition: loss averaged over all cities and per city losses. The solid horizontal line shows the overall loss of the winner. The red pluses show the per-city label coverage in the core competition.}
}
{%
\subfigure[Core competition]{%
\label{fig:leaderboard_core}
\includegraphics[height=5.0cm]{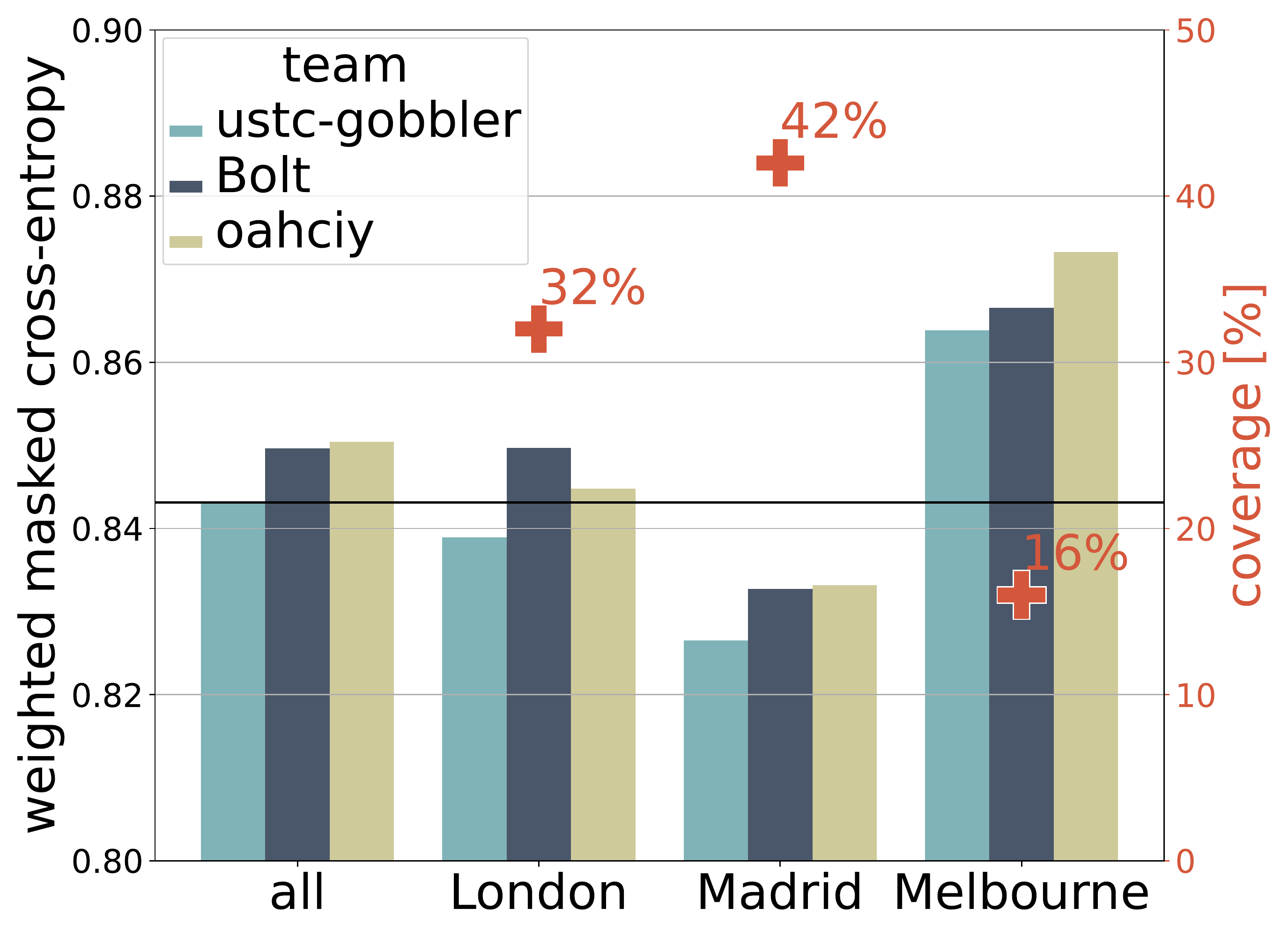}
}\qquad 
\subfigure[Extended competition]{%
\label{fig:leaderboard_extended}
\includegraphics[height=5.0cm]{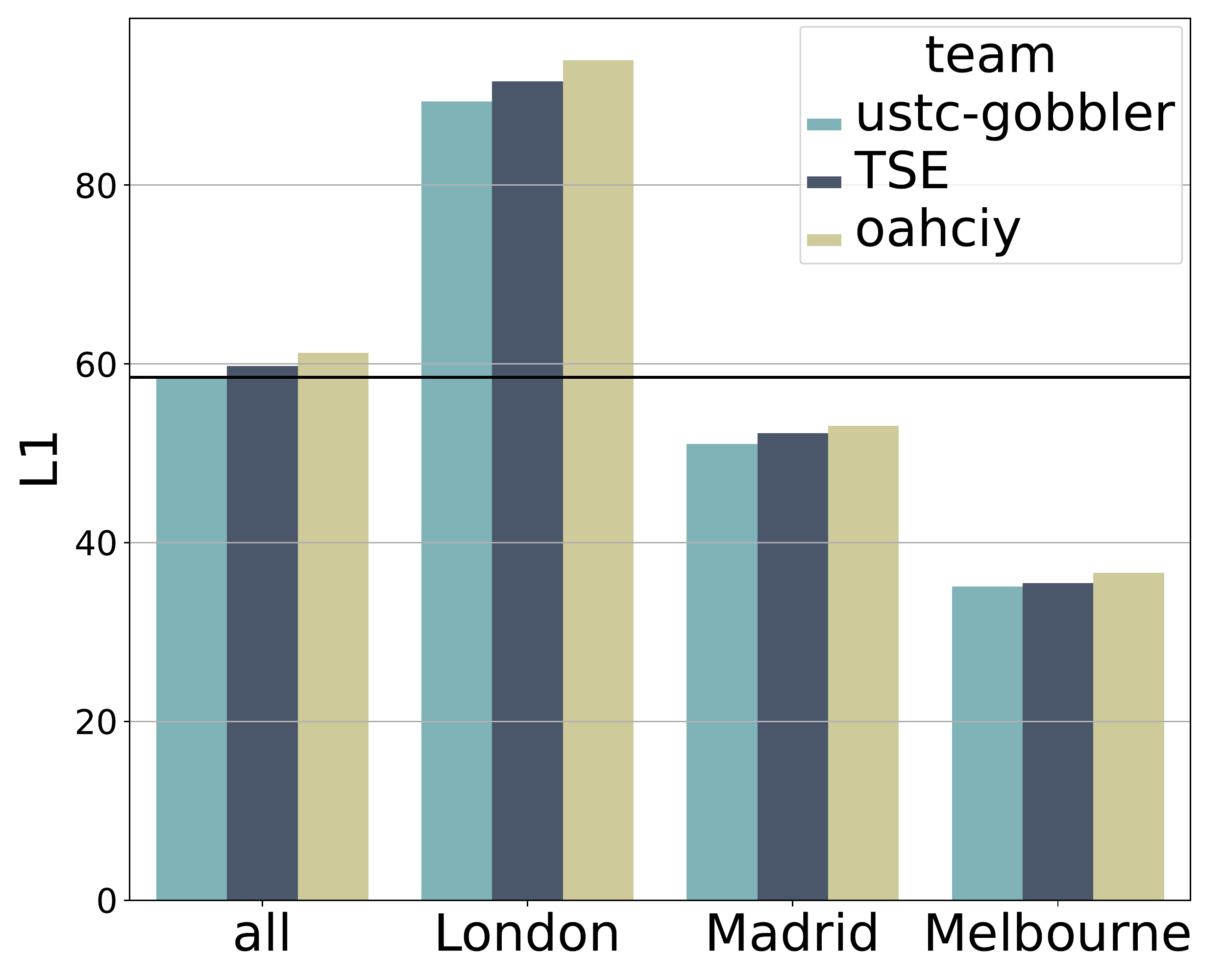}
}
}
\vspace*{-5mm}
\end{figure}

Figure~\ref{fig:leaderboard_core} shows the overall and per-city losses for the first 3 participants in the core competition.
We see the highest loss level in Melbourne and the lowest loss level in Madrid.
The loss level  shows an inverse correlation with the per-segment label coverage  (London: 32\%, Madrid: 42\%, Melbourne: 16\%) and a correlation with the imbalance of the label distribution. See Appendix~\ref{app:gt} for more details.

The loss levels in the extended competition are shown in 
Figure~\ref{fig:leaderboard_extended}.
They show an inverse correlation with the narrowness of the label distribution -- the Melbourne distribution with the highest peak and narrowest shape has the lowest loss level, whereas London with a flat peak and the fattest tail shows the highest loss level.
See Appendix~\ref{app:gt}.

\subsection{Do Models Learn a Historic Distribution?}\label{sec:historic_distribution}

\begin{figure}[htb]
  \centering
  \includegraphics[width=0.85\linewidth]{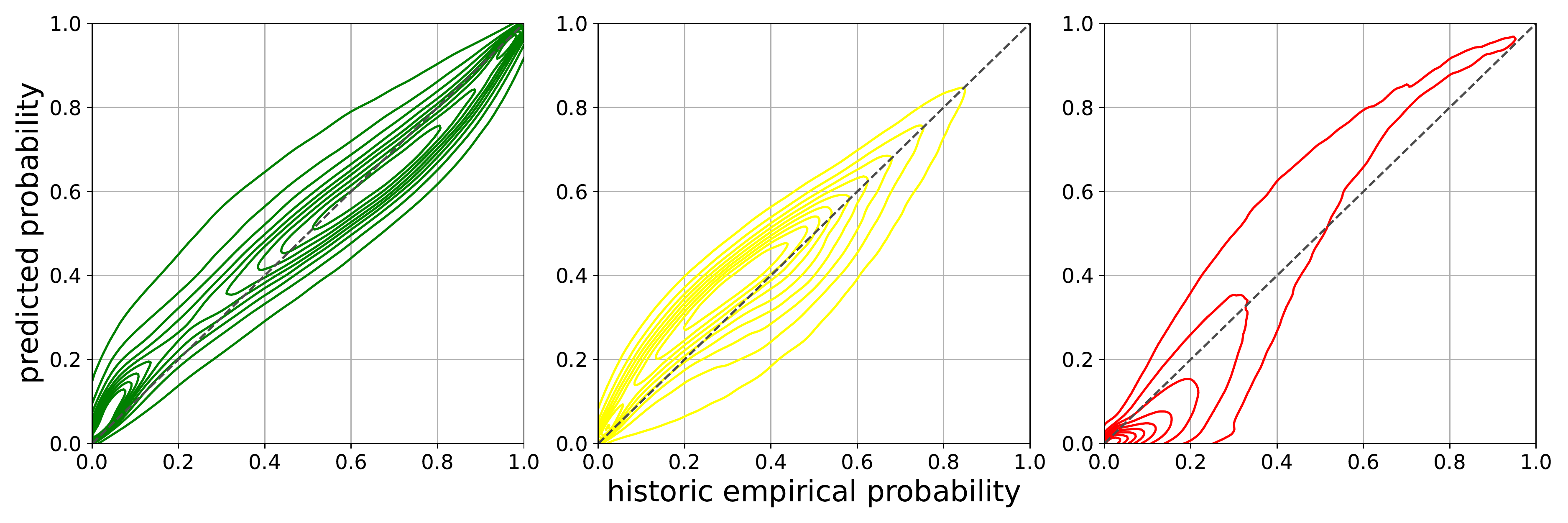}
  \vspace*{-6mm}
  \caption{Probabilites from re-weighted logits vs. historic ground-truth distribution, 
  submission by \cite{oahciy}, test set for London for 14:00--18:00. The historic empiric distribution is computed on the training data.} 
  \label{fig:kde}
  \vspace*{-5mm}
\end{figure}

In Figure~\ref{fig:kde}, we plot 
the edge-wise hourly historic empiric label distribution  (not taking into account day of week, only the hour of day) against the re-weighted model outputs
(submission by oahciy \cite{oahciy}, whole test set for London).
We see a strong correlation between the historic distribution and the predicted distribution after re-weighting. 
Details on the re-weighting and plots for other cities have a similar shape, see Appendix~\ref{app:kde}. 

\section{Summary and Outlook}

The \t4c Traffic Map Movie datasets offer a unique source of massive floating car data, unprecedented in its scale and availability across different metropolitan areas. The real-world data has been provided by HERE Technologies and the spatio-temporal aggregation is privacy-preserving. This has already enabled successful short-term traffic predictions \citep{Kreil_Traffic4cast_2019,Kopp_Traffic4cast_2020} as well as a demonstration of transfer learning in this context \citep{eichenberger_traffic4cast_2021}.

This year \t4c, for the first time, addressed directly application relevant challenges: taking as input only sparse traffic count data, can we for the whole city predict congestion classes, as we know them from the red-yellow-green maps in navigation apps, or even longer segment travel time (ETA)?
In this new setup, the \t4c traffic map movies provided the prediction labels by merging the aggregated speed-readings with a full road-graph. As a result also a general traffic segment speed graph dataset has already been made available for 10 cities \cite{mets-10}, and we see this development as a first step towards a traffic graph benchmark dataset in the ML domain.

The recurring nature of general traffic patterns allows for good prediction of common situations but special attention and treatment of rarer or more anomalous congestion events is also needed \citep{eichenberger_traffic4cast_2021}. We therefore used class weighting and super-segment sampling as discussed in Section~\ref{sec:metric}.

The spatial sparsity of the stationary vehicle detectors and their data being associated with nodes (and thus undirected) obviously limits the amount of local information that can be extracted. Nevertheless, just from the counts input, the models seem to capture the historic distribution well.
Interestingly, this made the task amenable to both GBM and GNN.
Furthermore, a more imbalanced label distribution and lower coverage made the core competition more difficult in Melbourne than in London and in Madrid.

It would be interesting to use further input signals and even to reverse the task and predict the ground-truth like stationary vehicle detector information from the more fine-grained GPS data. In a similar vein, this could also inform better placement of vehicle detectors for an improved detection of the global traffic state.

\appendix
\acks{We would like to thank HERE technologies for making our competition data available.}
\section*{Author Contributions Statement}

\addcontentsline{toc}{section}{Author Contributions Statement}
Following CRediT (Contributor Roles Taxonomy\footnote{\url{https://credit.niso.org/}}), the authors have contributed as follows.
\textit{Writing -- original draft, Software}: M.N., Ch.E.
\textit{Data Curation}: M.N., Ch.E, K.M.
\textit{Software (baselines)}: M.N., Ch.E., H.M., M.S., R.S., D.S.
\textit{Methodology (competition)}: L.D., C.W, D.L., M.Z., M.L., A.I., X.W., C.L., Q.L., V.M., Y.L., J.L, J.L., Y.G., F.G., J.M., Y.W., H.H., H.F.
\textit{Conceptualization}: M.N., Ch.E, F.T., M.K., D.K, S.H.
\textit{Writing -- review \& editing}: all.

\addcontentsline{toc}{part}{References}
\bibliography{references.bib, TrafficForecast.bib}

\newpage

\section{More Details on Stationary Vehicle Detector Data}\label{app:loop_counters}

We provide normalized counts with id, lat, lon, heading, day, 96 volume counts per day, using NaN for no data.
Madrid and Melbourne are providing \SI{15}{\minute} aggregates. We are normalizing the \SI{5}{\minute} sliding windows in London to the same \SI{15}{\minute} time bins for convenience and consistency. The data in Madrid contains the heading for each loop counter, whereas for London and Melbourne this is not readily available and could be derived from the associated road segment via parsing street designation; hence, we decide to assign the counter data to nodes without using the heading information.

\paragraph{Snapping}
As seen in  Table~\ref{tab:loop_counters_new},  multiple detectors can be close to the same road graph node. In this case, their counts are added up. The reference  to the city-specific counter IDs is tracked in the list in the \verb+counter_info+ attribute for road graph nodes.
\paragraph{Splitting}
For Melbourne,  each detector can be composed of up to 24 individual detectors as detailed in \cite{melbourne_traffic_signal_configuration_data_sheets}. Unfortunately, these documents are  not readily parsable. Hence, we take the sum of the individual detector readings  and use only the common prefix as ID. 
This can lead to the situation that  we need to split the detector value among multiple road graph nodes number of road graph nodes at large junctions in Melbourne. The number of nodes the value is  split into is tracked in the \verb+num_assigned+ attribute.

\paragraph{}
The data specification can be found in our competition GitHub repo, and code for downloading and normalizing vehicle detector data can be found in the GitHub repo for \cite{mets-10}.

\section{More Details on Road Graph}\label{app:road_graph}
We use the road graph derivation \texttt{OSMnx} \cite{BOEING2017126} as described in \cite{mets-10} with the following additions:
\begin{itemize}
    \item \textbf{introduce nodes for vehicle detector locations} We assign the counter to the nearest node in the road graph if the node is closer than \SI{40}{\meter} to the counter location. Else we
project the counter location to the nearest edge (discarding the counter if the distance to the nearest edge is greater than \SI{20}{\meter});
then, we split the nearest edge if the projection point is not close enough to either the start or end point of the edge. When
an edge is split, attributes like length, travel times and geometry need to split proportionally as well. Note that an edge in the
original OpenStreetMap road graph may be split multiple times in this approach.
    \item \textbf{filter out road segments with low volume}
    We use a $495\times436\times4$ volume heatmap per cell and heading; the heatmap is created as average of daily volumes from 30 randomly selected days in the training data.
    \item \textbf{remove unconnected components and dead-ends from the graph}
        \begin{itemize}
            \item \textbf{clean edges with no access} We remove edges with restricted access, such as `no', `private', `official', `permit', `delivery', `designated', `emergency'
            \item \textbf{clean edges with low volume} We then remove residential and unclassified edges with low volume (highest daily volume as per heatmap below 10 among all intersecting cells, of both directions if not one-way). Here, in order to avoid deleting \eg{} roundabouts upfront, we do not delete edges shorter than \SI{50}{\meter}  and keep nodes with a vehicle counter. 
            \item \textbf{clean dead end edges} We then delete dead-end edges.
            \item \textbf{clean isolates} and remove isolates (\ie{} nodes without neighbors).
            \item \textbf{clean self loops + clean isolates + clean dead end edges} Furthermore, we filter self-loops shorter than \SI{300}{\meter}. As this can introduce new isolates and dead end edges, we remove those newly introduced.
            \item \textbf{clean no neighbors}
            \item \textbf{clean sub graphs + clean dead end edges} We  keep only the largest connected component of the graph. 
            \item \textbf{clean circle ramps + clean end circles + clean isolates} We remove clean short circle ramps (\ie{} one node only connected to two other nodes which already have a direct connection).
            \item \textbf{clean multi edges} Finally, in order to avoid multi-edges in the competition, we split all multi-edges but the shortest one.
        \end{itemize}
\end{itemize}

\section{More Details on Super-Segment Sampling}\label{app:super_segments}

\begin{figure}[htb]
  \centering
  \includegraphics[width=0.85\linewidth]{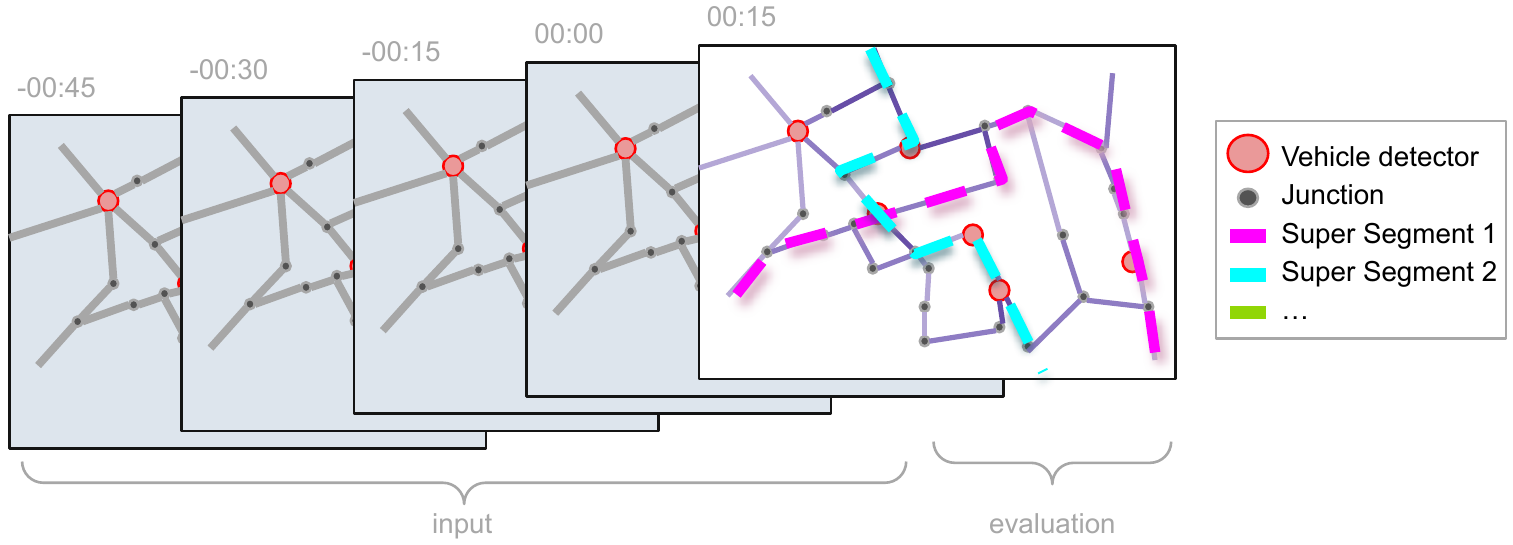}
  \vspace*{-5mm}
  \caption{\t4c 2022 Extended Task. Given one hour of sparse vehicle count data only (red disks) the task is to predict the average travel times along sampled super-segments in the city road-graph.} 
  \label{fig:evolution_extended_task}
  \vspace*{-5mm}
\end{figure}

For the extended task (see Figure \ref{fig:evolution_extended_task}) we are providing a choice of super-segments \citep{derrow-pinion_eta_2021} represented as lists of segment IDs. The super-segments for \t4c 2022 are sampled from the full road graph with the aim to capture corridors used by most typical routes on the road graph. By construction, there is at most one super-segment between any pair of nodes.

We start by choosing 400 {\bf connected key intersections}, which are those nodes with the highest maximum daily volume on any edge connected to it (incoming or outgoing), weighted by OSM importance, and which have at least 3 neighbor nodes in the graph.
We then go through this candidate list, removing neighbors of chosen nodes from the candidate list. For each city, there is also a hand-curated whitelist of nodes which is added to this candidate list.

From these filtered connected key intersections, we first derive a list of {\bf super-segments} for each of these key intersections.
For each connected key intersection, we look at the other connected key intersections in increasing circles defined by radius of  beeline distance, length of the shortest path in meters and number of segments in the super-segment candidate. Key intersections reached will not be considered again for the same source key intersection. 

For the shortest path, Dijkstra algorithm is used with edge weights linear in the length of the edge (in meters) and inversely proportional to the edge importance:
$weight(ed) = ((6.0 - importance(ed)) / 2) * length(ed)$, penalizing longer and unimportant edges. The search for super-segments from a connected key intersection will be aborted once a super-segment further away than \SI{10000}{\meter} is found or more than 3 super-segments have been found for a source and the search will be aborted after the 4th search circle.

For each city, there is also a hand-curated whitelist of super-segments which is added to the list of sampled super-segments.

\section{More Details on Derivation of Output Labels}\label{app:output_labels}


We refer to \cite{mets-10} for the derivation of dynamic edge speeds and edge free flow speeds. Here, we only describe the derivation of labels from there on.

\subsection{Congestion Classes}

To the best of our knowledge, there are only informal descriptions of how tech companies derive the congestion classes in their online maps, \eg \cite{color-in-google-maps,uber-movement-faqs,your-favorite-traffic-map-is-lying-to-you}.
Hence, we choose one of them, namely \cite{mapping-traffic-congestion},
and derive the congestion classes in the following way:
\begin{verbatim}
def extract_cc(segment_speed_stats, t, freeflow_speed_kph):
    median_speed = segment_speed_stats.median_speeds[t]
    if median_speed == 0:
        return 0
    if median_speed == 255:
        return 0
    assert(freeflow_speed_kph > 0)
    probe_volume = segment_speed_stats.volumes[t]
    median_speed_kph = segment_speed_stats.median_speeds_kph[t]
    congestion_factor = median_speed_kph / freeflow_speed_kph
    if congestion_factor < 0.4 and probe_volume >= 5:
        return 3
    elif congestion_factor >= 0.4 and \ 
         congestion_factor < 0.8 and probe_volume >= 3:
        return 2
    elif congestion_factor >= 0.8 and probe_volume > 0:
        return 1
    else:
        return 0
\end{verbatim}        
The congestion factor reflects the percentage of free flow speed. 
If the median speed in the data is 0 or 255, we do not classify it as this hints at data corruption. 
We set 0.4 and 0.8 as boundaries to classify for congestion (red, 3), reduced (yellow,2), and uncongested (green,1). In addition, we impose volume limits for the classification, requiring more evidence to classify as red than for yellow.
For US highways (speed limit \SI{65}{mph}), green is moving at least \SI{50}{mph} (congestion level 80\%), orange approximately 25--50 miles per hour (congestion level 40--80\%), red below 25 (congestion level 40\%).

\subsection{Super-Segment ETAs}

ETAs are derived for all super-segments with a two-level defaulting mechanism: for each edge $ed$ in the super-segment, a speed $v(ed)$ is derived, taking the current median speed from the dynamic data as first priority, free flow speed as second priority and signalled maxspeed as third priority. The speed is clipped below at $0.5 \frac{km}{h}$.
If ETA is longer than the 15 min slot and half of both neighbor time slots, use 30 minutes plus the mean speed of all three slots.
The eta for super-segment with $edges$ is then derived by $\sum_{ed \in edges} \frac{length(ed)}{v(ed)}$:

\begin{verbatim}
# prec-computed for whole day
def compute_edge_speeds_for_one_day(...):
    for uv, maxspeed in edge_maxspeeds_kph.items():
        if uv in edge_free_flows_kph:
            free_flow = edge_free_flows_kph[uv]
            speeds = [free_flow for _ in range(96)]
            sources = ["free_flow" for _ in range(96)]
        else:
            speeds = [maxspeed for _ in range(96)]
            sources = ["maxspeed" for _ in range(96)]
            maxspeed_cnt += 1
        edge_speeds[uv] = {"speeds": speeds, "sources": sources}
    print(f"{maxspeed_cnt} / {len(edge_speeds)} edges only have maxspeed")
    for t in range(0, 96):
        tsc_df = sc_df[sc_df["t"] == t]
        for u, v, ms in zip(tsc_df["u"], 
                            tsc_df["v"], 
                            tsc_df["median_speed_kph"]):
            esp = edge_speeds[(u, v)]
            esp["speeds"][t] = ms
            esp["sources"][t] = "current"
    

def compute_eta(edges, edge_speeds, t):
    path_eta_s = 0

    for ed in edges:
        e = ed["edge"]
        esp = edge_speeds[e]
        edge_speed_kph = esp["speeds"][t]
        speed_used = esp["sources"][t]

        length_m = ed["length"]
        
        if edge_speed_kph < 0.5:
            edge_speed_kph = 0.5
        edge_speed_mps = edge_speed_kph / 3.6
        edge_eta_s = length_m / edge_speed_mps

        if edge_eta_s > 1800:
            # If ETA longer than the 15 min slot and half of 
            # both neighbor time slots,
            # use 30 minutes plus the mean speed of all three slots.
            speeds = esp["speeds"][max(t - 1, 0) : t + 2]
            sources = esp["sources"][max(t - 1, 0) : t + 2]
            edge_speed_kph = mean(speeds)
            edge_speed_mps = edge_speed_kph / 3.6
            edge_eta_s = 1800 + (length_m / edge_speed_mps)

        path_eta_s += edge_eta_s

    return path_eta_s
\end{verbatim}

Note that, upon congestion, there will be a higher chance of having data going into current median speeds, and in those cases the second and third line will be appropriate; the clipping is necessary since, in case of congestion, the edge speed can get very low as slow or standing vehicles will emit many more data points along the same edge.
Furthermore, in a city environment (in contrast to a motorway locked down by a traffic), a full traffic breakdown is also less likely; as the ETAs are computed for full \SI{15}{\minute} intervals, the ETA reflects on an expectation. 

Such a heuristic approach is also behind industrial ETA routing machines -- all companies basing their ETA products on floating car data will not have a full view of traffic and will have default values.

\section{Ground Truth Label Distribution Core and Extended Competition}\label{app:gt}

\subsection{Core competition label and coverage distribution}
\begin{figure}[htb]
  \centering
  \includegraphics[width=0.95\textwidth]{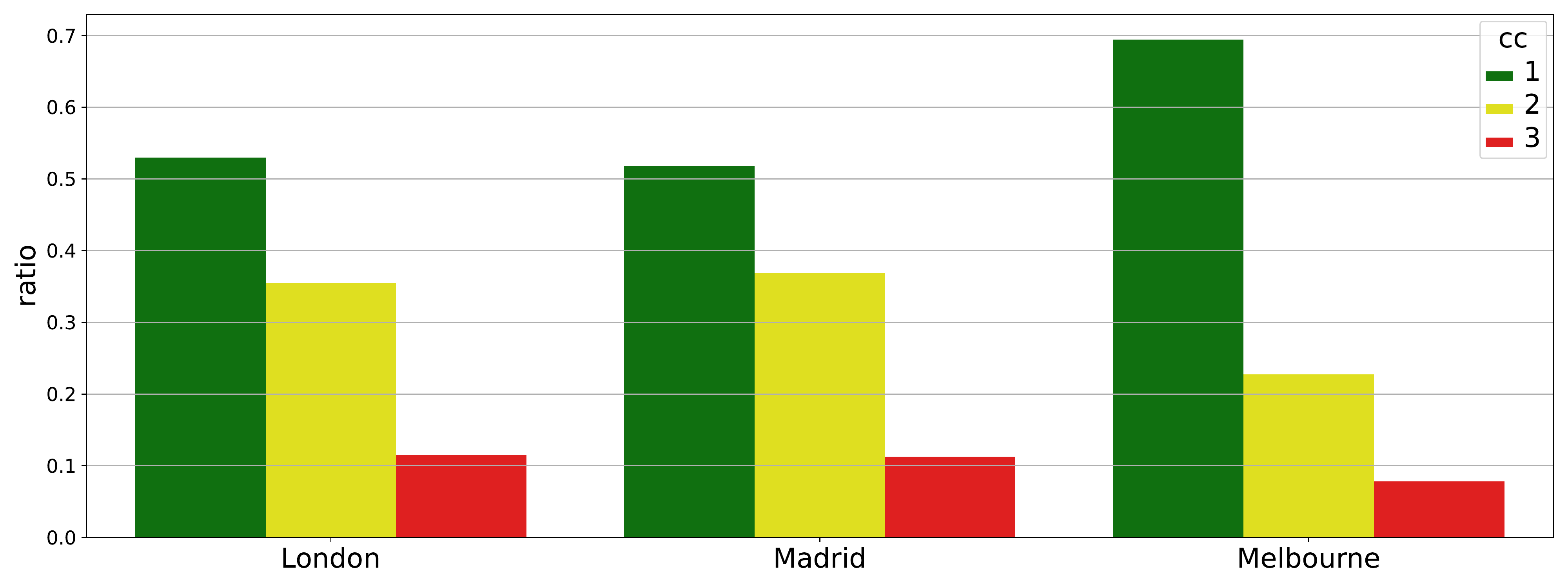}
  \vspace*{-2mm}
  \caption{Label distribution in the test set of the core competition for the three cities London, Madrid, Melbourne.} 
  \label{fig:ev06_ground_truth_distribution_core_label_ratio}
  \vspace*{-2mm}
\end{figure}

\begin{figure}[htb]
  \centering
  \includegraphics[width=0.95\textwidth]{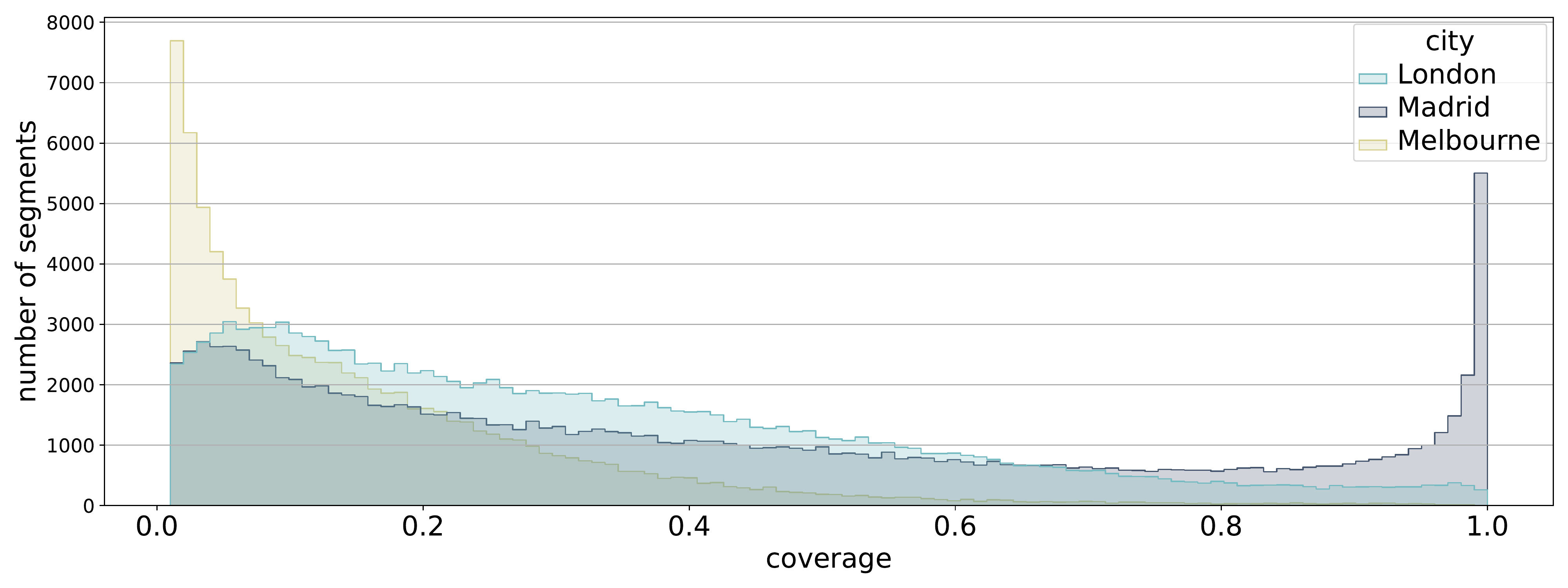}
  \vspace*{-2mm}
  \caption{Distribution of per-segment label coverage in the test set of the core competition for the three cities London, Madrid, Melbourne.} 
  \label{fig:ev06_ground_truth_distribution_core_coverage_histogram}
  \vspace*{-2mm}
\end{figure}

Figure~\ref{fig:ev06_ground_truth_distribution_core_label_ratio} shows that the labels are most equally distributed in Madrid, and most imbalanced in Melbourne.
Figure~\ref{fig:leaderboard_core} shows that the the per-segment label coverage is highest in Madrid and lowest in Melbourne.
In addition, Figure~\ref{fig:ev06_ground_truth_distribution_core_coverage_histogram} shows that Madrid has many segmenst with a very high coverage, whereas in Melbourne there are almost none with a very high coverage.

\clearpage
\subsection{Extended competition label distribution}

\begin{figure}[htb]
  \centering
  \includegraphics[width=\textwidth]{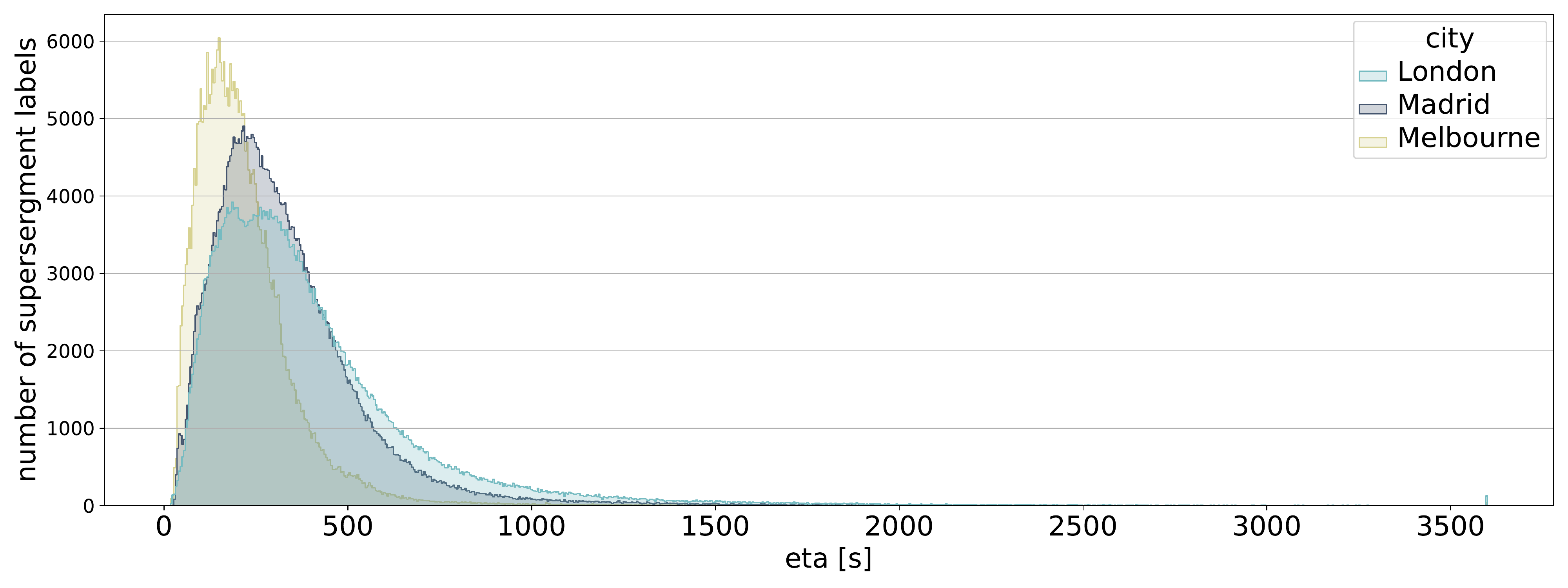}
  \vspace*{-2mm}
  \caption{Histogram of super-segment labels in the test set of the extended competition for the three cities London, Madrid, Melbourne.} 
  \label{fig:ev06_ground_truth_distribution_extended_hist}
  \vspace*{-2mm}
\end{figure}

Figure~\ref{fig:ev06_ground_truth_distribution_extended_hist} shows that London has the longest tail in super-segment ETAs, and Melbourne has the most narrow distribution.
\clearpage

\section{Notation Weighted Masked Cross-Entropy Loss}\label{app:metrics}
Following the notation of \cite{torch.nn.CrossEntropyLoss},
\begin{equation}
\ell(\hat{y}, y) = \frac{1}{k} \cdot \sum_{n=1}^N   l_n, \quad
          l_n = - w_{y_n} \log \frac{ \exp(\hat{y}_{n,y_n})}{\Sigma_{c=1}^{C} \exp(\hat{y}_{n,c})} \cdot \mathbbm{1}\{y_n \not= \kappa\}, \quad k=\sum_{n=1}^N   w_{y_n} \cdot \mathbbm{1} \{y_n \not= \kappa\} \label{eq:weightedmaskedCE}
\end{equation}
where
$\hat{y} \in \mathbb{R}^{N \times C}$ are the input logits, 
$y \in \{0,...,C\}^N$ is the target,
$w \in \mathbb{R}^C$ is the class weight,
$C \in \mathbb{N}$ is the number of classes to be predicted,
$N \in \mathbb{N}$ is the number of samples,
$\kappa$ specifies a target class that is ignored (\ie models must not predict no data).
Cross-entropy loss will penalize small predicted probabilities disproportionately.

In our setting, $C=4$ and $\kappa=0$, \ie{}, we always mask on unclassified edges in the ground truth (0 = unclassified, 1 = green (uncongested), 2 = yellow (slowed down/warning), 3 = red (congested)), $N$ goes over edges and timestamps.

In problems where infrequent classes are nonetheless important, macro-averaging may be a means of highlighting their performance \cite{scikit-learn_model_evaluation}.  In our case, since we have more red than yellow than green in all cities and the congested situations are particularly important.
We compute the macro-averaged  class weights computed on the $\tilde{N}$ training labels $\tilde{y}$, 
\begin{equation}
w_c = \frac{\tilde{N}}{|C| \cdot \Sigma_{n=1}^{\tilde{N}}{   \mathbbm{1} \{\tilde{y}_n = c\} }},\quad c \in C-\{\kappa\},
\end{equation}
giving equal total weight to each class as can be seen from the decomposition
$$
k = \sum_c N_c \cdot w_c = \sum_c  \frac{N_c \cdot \tilde{N}}{ |C| \cdot \tilde{N}_c} \approx \sum_c \frac{N_c \cdot {N}}{ |C| \cdot {N}_c} = \sum_c \frac{N}{|C|}
$$
if we assume assume the label fractions to be almost the same in the test and training set, \ie if we assume $\frac{N_c}{N}\approx\frac{\tilde{N}_c}{\tilde{N}}$, 
where  $N_c=\sum_{i=1}^N{   \mathbbm{1} \{y_i = c\} }$ the number of labels of class $c$ in the test set and where  $\tilde{N}_c=\sum_{i=1}^{\tilde{N}}   \mathbbm{1} \{\tilde{y}_i = c\}$ is the number of labels of class $c$ in the training labels $\tilde{y}$ from which the weights $w_c$ were computed.

\section{Loss Distribution Core per Ground-Truth Label}\label{app:loss_gt}

\begin{figure}[htb]
  \centering
  \includegraphics[width=\textwidth]{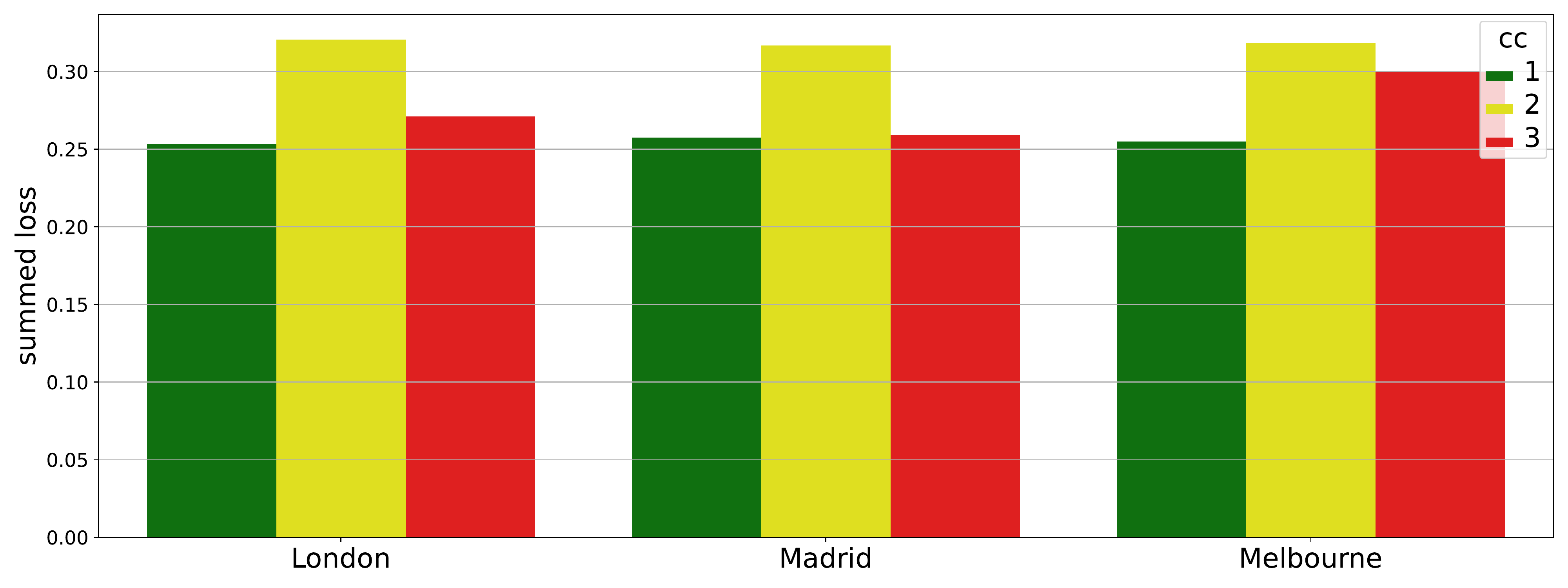}
  \vspace*{-2mm}
  \caption{Summed loss per ground-truth in the test set of the core competition for the three cities London, Madrid, Melbourne for team oahciy. The sum of the green, yellow and red bar per city gives the per-city losses of Figure~\ref{fig:leaderboard_core}.} 
  \label{fig:ev08_sum_loss_by_gt_oahciy}
  \vspace*{-2mm}
\end{figure}

\begin{figure}[htb]
  \centering
  \includegraphics[width=\textwidth]{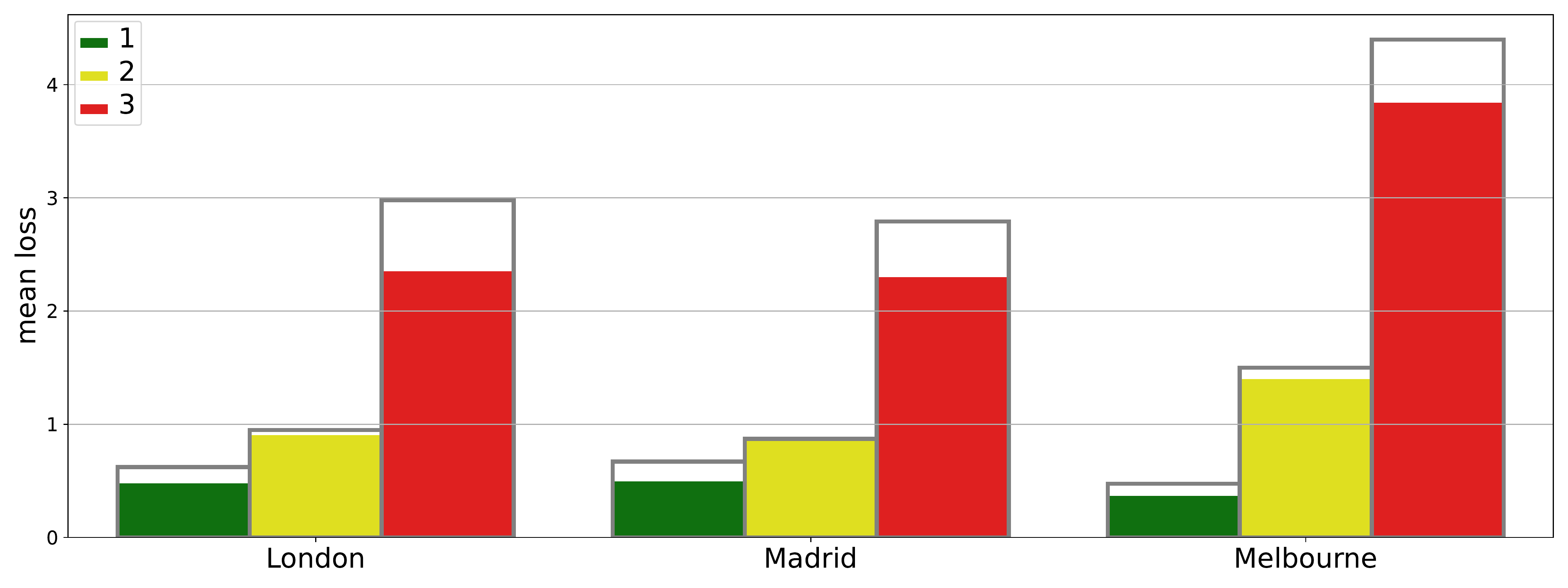}
  \vspace*{-2mm}
  \caption{Mean loss per ground-truth in the test set of the core competition for the three cities London, Madrid, Melbourne for team oahciy. The sum of the green, yellow and red bar per city gives the per-city losses of Figure~\ref{fig:leaderboard_core}. The grey boxes reflect the class weights.} 
  \label{fig:ev08_mean_loss_by_gt_oahciy}
  \vspace*{-2mm}
\end{figure}

Figure~\ref{fig:ev08_sum_loss_by_gt_oahciy} shows that predicting green and yellow was similarly difficult in all three cities, whereas predicting red was harder in Melbourne and easiest in Madrid.
This is strongly inversely correlated with the label distribution of Figure~\ref{fig:ev06_ground_truth_distribution_core_label_ratio}.
The city-wise mean and summed losses per ground truth class $c$ were computed as follows.
Filtering out the ignored class in Equation~(\ref{eq:weightedmaskedCE}), we have 
$$k=\sum_{n=1}^N w_{y_n} = \sum_c N_c \cdot w_c = \sum_c  N_c \frac{\tilde{N}}{|C| \cdot \tilde{N}_c} \approx N.$$
Then, we consider
$$summedloss(c) = \frac{1}{k} \cdot \sum_n  l_n \cdot \mathbbm{1} \{y_n = c\},\quad meanloss(c) = \frac{1}{N_c} \sum_n l_n \cdot \mathbbm{1} \{y_n = c\}.$$
The mean losses of Figure~\ref{fig:ev08_mean_loss_by_gt_oahciy} reflect the summed losses scaled by the class weights as we have 
$$
meanloss(c)/ (w_c \cdot |C|)  =  
\frac{|C| \cdot N_c}{N_c \cdot N \cdot |C|} \cdot \sum_n  l_n \cdot \mathbbm{1} \{y_n = c\} 
\approx \frac{1}{k} \cdot \sum_n  l_n \cdot \mathbbm{1} \{y_n = c\} = summedloss(c).
$$



\begin{figure}[htb]
  \centering
  \includegraphics[width=\textwidth]{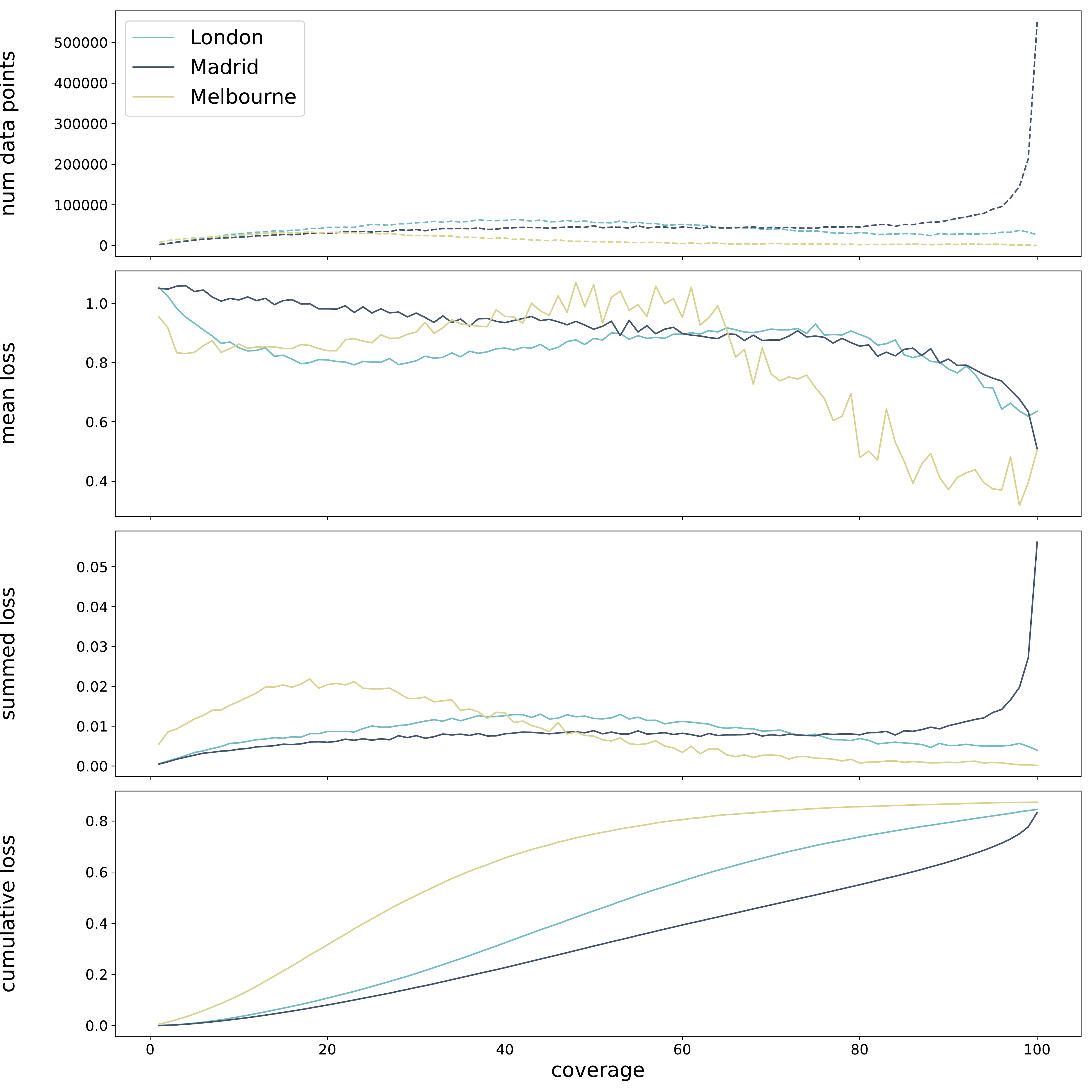}
  \vspace*{-2mm}
  \caption{Number of data points and losses per coverage (mean per data point, summed per coverage, and cumulative).} 
  \label{fig:ev08_loss_by_coverage_oahciy}
  \vspace*{-2mm}
\end{figure}

\clearpage
Figure~\ref{fig:ev08_loss_by_coverage_oahciy} shows losses binned by per-segment coverage. 
The mean loss curve for Madrid is almost monotone decreasing, whereas London sees a slightly elevated plateau around 70\% coverage and Melbourne sees a more accentuated plateau around in the are 50--60\%; the mean loss curve reflects to traffic mix in those streets.
Due to the steep decrease in mean loss in Madrid for higher coverage, the high number of data points for high coverage still results in an overall loss as seen in the cumulative plot.

\clearpage

\section{More on Capturing the Historic Distribution}\label{app:kde}

Here, we detail on the comparison of model outputs with the hourly, segment-wise historic distribution as discussed in Section~\ref{sec:historic_distribution}
The per-city class weights from Section~\ref{app:metrics} will drive the models to overestimate the rare classes (red and yellow) and underestimate the non-rare classes (green), as shown by the plots without re-weighting.
The re-weighted probabilities are derived as
\begin{equation}
\hat{p}_{n,c} = b_{n} \cdot \exp(\hat{y}_{n,c} \cdot w_{c})
\end{equation}
from the model outputs $\hat{y}_{n,c}$ and normalization factor $b_n$.

\subsection{London}
%
%
%
\begin{figure}[htb]
  \centering
  \includegraphics[width=0.95\linewidth]{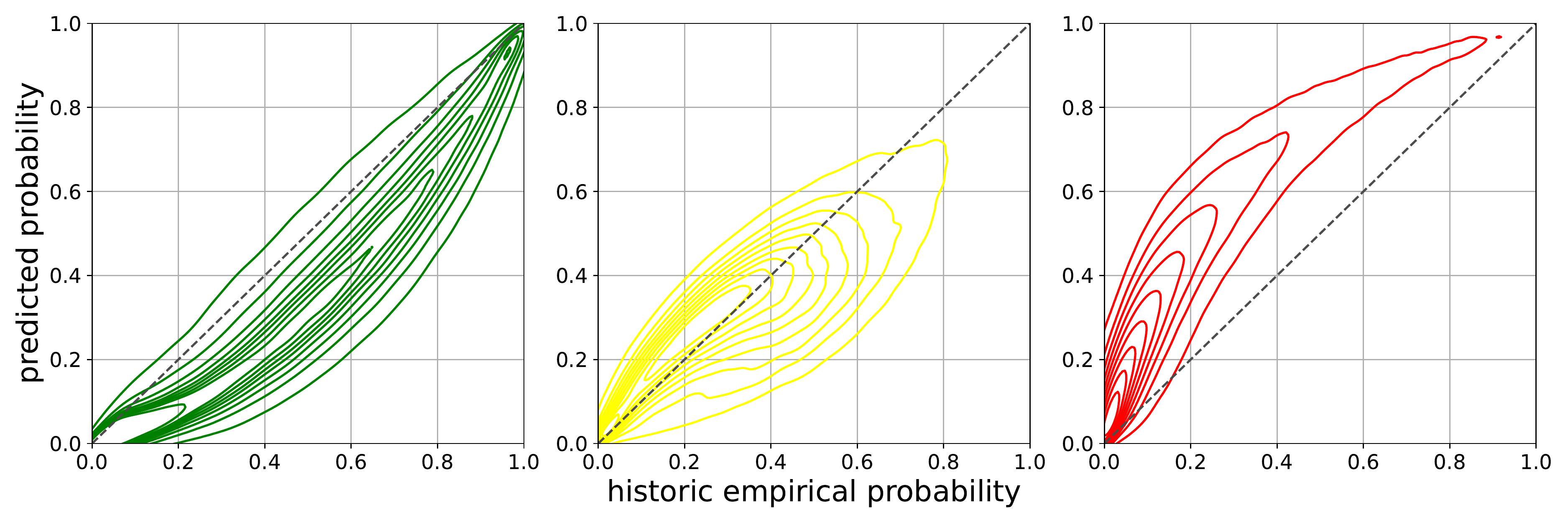}
  \vspace*{-2mm}
  \caption{Probabilites from  logits without re-weighting vs. historic ground-truth distribution, 
  submission by \cite{oahciy}, test set for London for 14:00--18:00. The historic empiric distribution is computed on the training data.} 
  \label{fig:ev07_score_surprise_london_oahciy_kde_exp_surprise_prediction_exp_surprise_gt}
  \vspace*{-2mm}`
\end{figure}
\clearpage

\subsection{Madrid}
%
%
\begin{figure}[htb]
  \centering
  \includegraphics[width=0.95\linewidth]{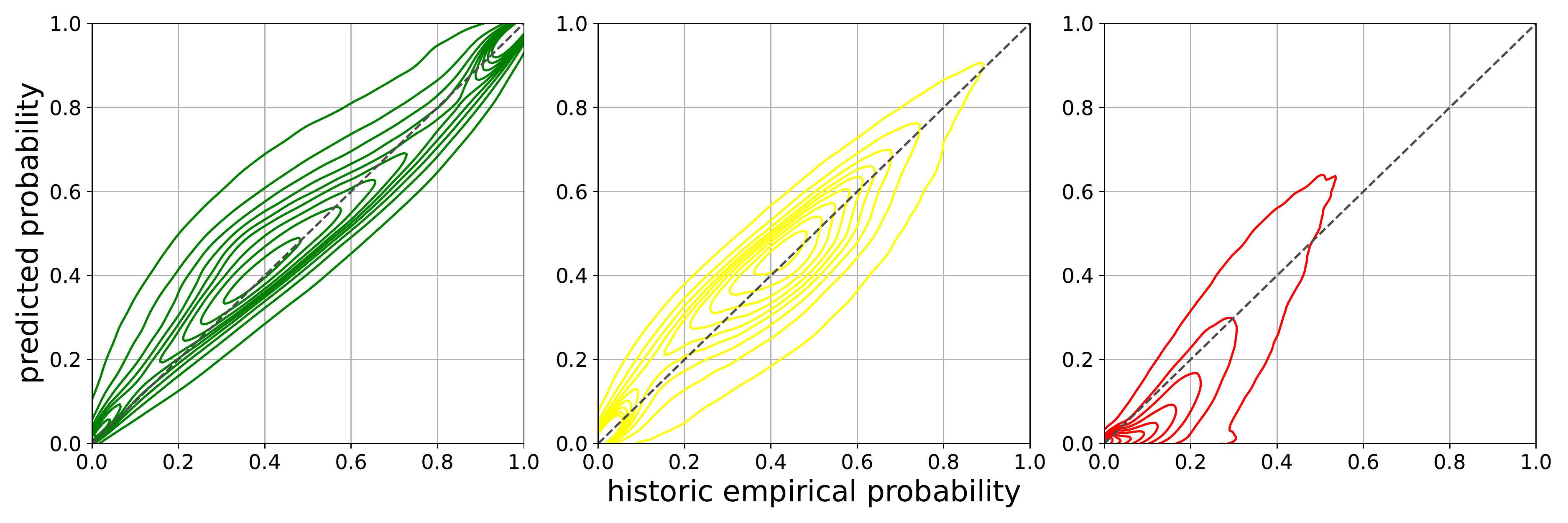}
  \vspace*{-2mm}
  \caption{Probabilites from re-weighted logits vs. historic ground-truth distribution, 
  submission by \cite{oahciy}, test set for Madrid for 14:00--18:00. The historic empiric distribution is computed on the training data.} 
  \label{fig:ev07_score_surprise_madrid_oahciy_kde_exp_surprise_prediction_weight_corrected_exp_surprise_gt}
  \vspace*{-2mm}
\end{figure}

\begin{figure}[htb]
  \centering
  \includegraphics[width=0.95\linewidth]{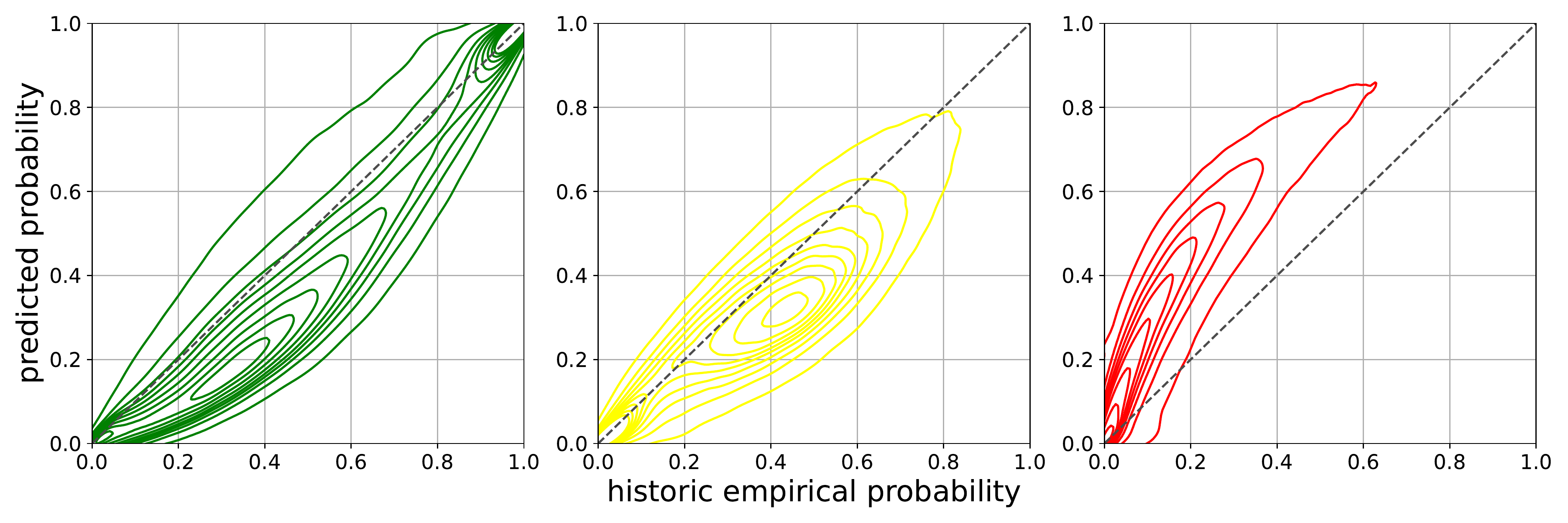}
  \vspace*{-2mm}
  \caption{Probabilites from  logits without re-weighting vs. historic ground-truth distribution, 
  submission by \cite{oahciy}, test set for Madrid for 14:00--18:00. The historic empiric distribution is computed on the training data.} 
  \label{fig:ev07_score_surprise_madrid_oahciy_kde_exp_surprise_prediction_exp_surprise_gt}
  \vspace*{-2mm}`
\end{figure}
\clearpage

\subsection{Melbourne}
%
%
\begin{figure}[htb]
  \centering
  \includegraphics[width=0.95\linewidth]{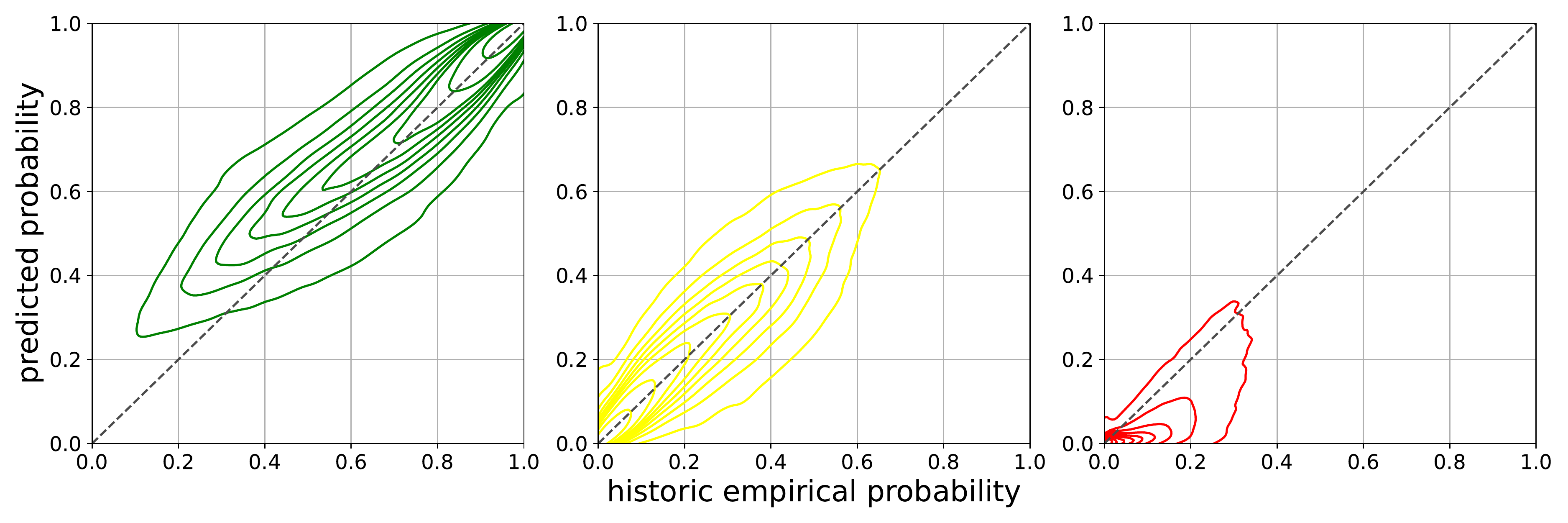}
  \vspace*{-2mm}
  \caption{Probabilites from re-weighted logits vs. historic ground-truth distribution, 
  submission by \cite{oahciy}, test set for Melbourne for 14:00--18:00. The historic empiric distribution is computed on the training data.} 
  \label{fig:ev07_score_surprise_melbourne_oahciy_kde_exp_surprise_prediction_weight_corrected_exp_surprise_gt}
  \vspace*{-2mm}
\end{figure}

\begin{figure}[htb]
  \centering
  \includegraphics[width=0.95\linewidth]{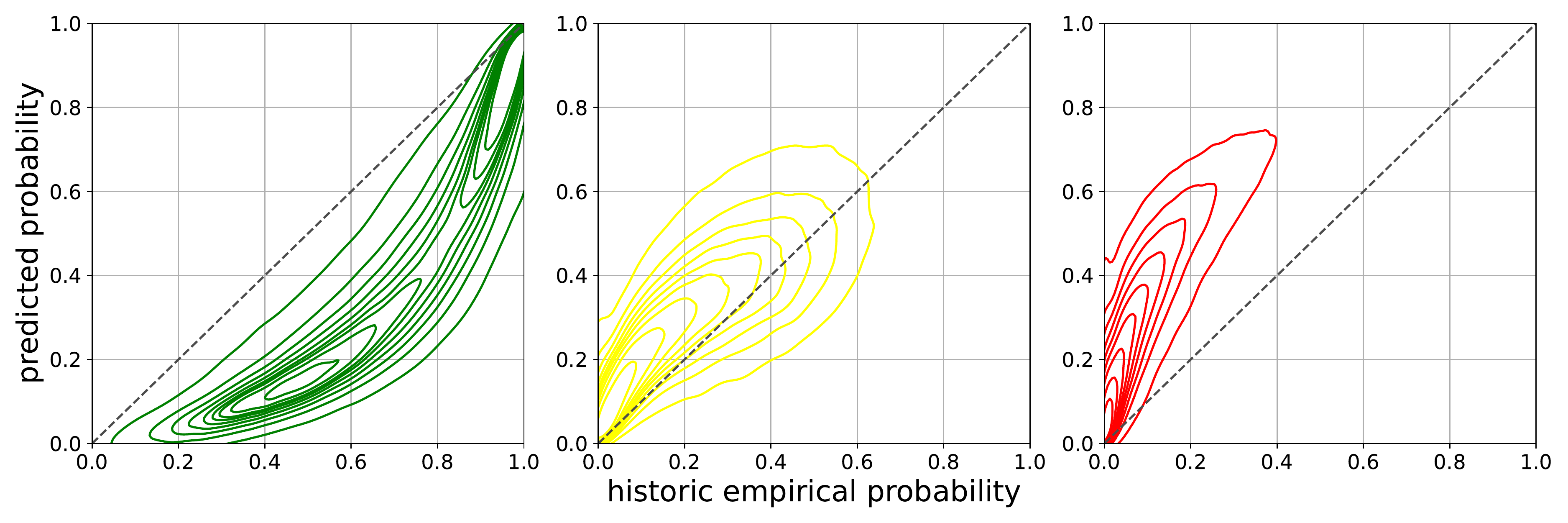}
  \vspace*{-2mm}
  \caption{Probabilites from  logits without re-weighting vs. historic ground-truth distribution, 
  submission by \cite{oahciy}, test set for Melbourne for 14:00--18:00. The historic empiric distribution is computed on the training data.} 
  \label{fig:ev07_score_surprise_melbourne_oahciy_kde_exp_surprise_prediction_exp_surprise_gt}
  \vspace*{-2mm}`
\end{figure}

\end{document}